# A Novel A.I Enhanced Reservoir Characterization with a Combined Mixture of Experts - NVIDIA Modulus based Physics Informed Neural Operator Forward Model

Clement Etienam*[1], Yang Juntao[1], Issam Said[1,] Oleg Ovcharenko[1], Kaustubh Tangsali[1,] Pavel Dimitrov[1], Ken Hester[1]

**ABSTRACT**

We have developed an advanced workflow for reservoir characterization that elegantly addresses the challenges of reservoir history matching. This novel approach integrates a **P**hysics **I**nformed **N**eural **O**perator (*PINO*) as a forward model within a sophisticated mixture of expert's framework, named **C**luster **C**lassify **R**egress (*CCR*). The process of inverse modeling is facilitated through an adaptive Regularized Ensemble Kalman Inversion (*a*REKI), which is particularly efficient for rapid uncertainty quantification tasks involved in reservoir history matching. In our methodology, the unknown fields of permeability and porosity are parameterized, capturing non-Gaussian posterior measures using advanced techniques such as a variational convolution autoencoder and the *CCR*. The *CCR* not only serves as exotic priors in our inverse modeling but also synergizes as a supervised model with the prior *PINO* surrogate to accurately simulate the nonlinear dynamics described by the Peaceman well equations in the forward model. The *CCR* approach is designed with flexibility, allowing for any distinct and independent machine learning algorithm to be applied across its three stages. The updating of the *PINO* reservoir surrogate is driven by a loss function derived from supervised data as well as the initial conditions and residuals of the governing black oil PDEs. The integrated model of the *PINO* and *CCR* now called ***PINO-Res-Sim*** outputs various crucial parameters, including pressures, water and gas saturations, and production rates for oil, water, and gas. This methodology has been validated against traditional numerical simulators in a controlled waterflooding experiment conducted on two synthetic reservoirs and a *water-alternating-gas* experiment of the *Norne* field, demonstrating remarkable accuracy. Further application of our *PINO-Res-Sim* surrogate in the *a*REKI history matching workflow showed its potential by efficiently recovering unknown permeability and porosity fields with a computational speedup of up to 100 - 6000 times faster than conventional simulation techniques. The learning phase for the *PINO-Res-Sim* model, conducted on an NVIDIA H100 with 80G of memory, was impressively efficient—taking approximately 30 minutes for 10,000 training samples in a synthetic 2D model, 5 hours in a 3D model, and 4 hours for 100 samples from the *Norne* field. This workflow is also compatible with ensemble-based methods, which are beneficial for sampling posterior densities in complex computational tasks where evaluating likelihoods is computationally expensive.

---

[1] NVIDIA
* Corresponding author

# 1 Introduction

## 1.1 Overview
Reservoir engineering is a critical field where professionals strive to develop dependable simulation models that can precisely emulate the intricate behaviours of subsurface reservoir systems. At the heart of this challenge lies the inverse problem, which involves using a set of observed measurements, denoted as $y$, to deduce a set of model parameters, $u$.

These observed data, direct functions of state variables, serve as a gateway to infer vital information about the model parameters amidst inherent uncertainties. Focused primarily on discrete inverse problems, i.e. systems characterized by a finite number of parameters, reservoir history matching emerges as a quintessential example. Here, the finite data, marred by measurement inaccuracies, render the exact estimation of reservoir model parameters from inconsistent and incomplete data via a forward model (likely flawed itself) a formidable task. Notwithstanding, pre-existing knowledge on the geological model, informed by cores, logs, and seismic data, alongside insights into the depositional environment, assists in navigating these complexities.

Reservoir history matching, aimed at aligning the reservoir model with actual historical dynamic production data to forecast future production and diminish uncertainty in reservoir descriptions, epitomizes an ill-posed problem. The discrepancy between the limited independent data and the multitude of variable combinations underscores the infinite possibilities of unknown reservoir properties capable of matching observed data. This dilemma is compounded by the inaccuracies and potential inconsistencies in reservoir information, leading to models riddled with uncertain parameters and, by extension, uncertain predictions.

The past decade has witnessed a growing emphasis on quantifying uncertainty in reservoir model predictions and descriptions to manage risk effectively. This fascination with uncertainty characterization has popularized the generation of multiple history-matched models. Yet, this approach does not necessarily provide an accurate assessment of uncertainty, a concept devoid of scientific significance outside the statistical and probabilistic domain. Bayesian statistics offers a coherent framework for addressing uncertainty, enabling the formulation of the posterior probability density function (pdf) for reservoir model parameters, conditioned on field measurements like production, electromagnetic, and seismic data.

Characterizing model uncertainty, therefore, boils down to sampling this posterior pdf, with a proper assessment of uncertainty in reservoir performance predictions emerging from statistical analysis of predicted outcomes across various models. Markov chain Monte Carlo (*MCMC*) methods, despite their theoretical appeal for sampling the posterior pdf, face challenges in high-dimensional scenarios due to the computational intensity required for evaluating the likelihood associated with each transition in state, rendering direct application to realistic reservoir problems impractical.

**Our Contribution:**
- We introduce a cutting-edge approach incorporating the Fourier neural operator surrogate model (*PINO*-surrogate), leveraging both data and physics loss within the NVIDIA Modulus framework to supplant traditional black oil model reservoir equations.
- This innovation paves the way for a novel inverse modelling workflow that utilizes the *PINO* surrogate in conjunction with exotic priors to meticulously calibrate the permeability and porosity fields of the target reservoir. Furthermore, this workflow is adeptly employed for forward uncertainty quantification and field optimization, showcasing its utility in enhancing reservoir characterization and modelling.
- Through our work, we aim to advance the integration of Fourier neural operator techniques in reservoir engineering, building upon the foundation of recent studies that have demonstrated the potential of such models in improving the accuracy and efficiency of reservoir characterization and predictive modelling processes.

## 1.2 Forward problem

### 1.2.1 Two Phase Flow
Our simplified model for two-phase flow in porous media for reservoir simulation is given as,

$$\varphi \frac{\partial S_w}{\partial t} - \nabla . [T_w(\nabla p_w + \rho_w gk)] = Q_w \; ; \; \varphi \frac{\partial S_o}{\partial t} - \nabla . [T_o(\nabla p_o + \rho_o gk)] = Q_o$$

Eqn. (1a)

where subscript $w$ stands for water and subscript $o$ stands for oil. The system is closed by adding two additional equations $P_{cwo} = p_o - p_w$ ; $S_w + S_o = 1$. This gives four unknowns oil pressure $p_o$, water pressure $p_w$, water saturation $S_w$ and oil saturation $S_o$. Gravity effects are considered by the terms $\rho_w gk$ and $\rho_o gk$. $\Omega \subset \mathbb{R}^n$ ($n = 2, 3$) is the modelling domain with boundary $\partial\Omega$, and $[0, t_f]$ is the time interval for which production data is available. $\varphi(x)$ stands for the porosity, and by $T_o, T_w$ and $T$ the transmissibilities, which are known functions of the absolute permeability $K$ and the water saturation $S_w$:

$$T_w = \frac{K(x)K_{rw}(S_w)}{\mu_w} \; ; T_o = \frac{K(x)K_{ro}(S_w)}{\mu_o} \; ; T = T_w + T_o$$

Eqn. (1b)

The relative permeabilities $K_{rw}(S_w)$ and $K_{ro}(S_w)$ are available as tabulated functions, and $\mu_w$ and $\mu_o$ denote the viscosities of each phase. $Q_o$, $Q_w$ and $Q = Q_o + Q_w$ define the oil flow, the water flow, and the total flow, respectively, which are measured at the well positions. Eqn. (1a) is solved with appropriate initial conditions and a no-flux boundary condition on $\partial\Omega$. We neglect the gravity terms $\rho_w gk$ and $\rho_o gk$, as well as capillary pressure (such that $P_w = P_o = P$). We set the following initial and boundary conditions $S_w(x, 0) = S_w^0(x)$ ; $P(x, 0) = P_w^0(x)$ ; $\nabla P . \nu = 0$. Here, $\nu$ is the outward unit normal to $\partial\Omega$, signifying no flux across the boundary. This yields the pressure and saturation equations as follows.

$$-\nabla . [T\nabla P] = Q$$

Eqn.(2a)

$$\varphi \frac{\partial S_w}{\partial t} - \nabla . [T_w(\nabla P)] = Q_w$$

Eqn.(2b)

### 1.2.2 Three Phase Flow
Our simplified model for three-phase flow in porous media for reservoir simulation is given as,

$$\varphi \frac{\partial S_w}{\partial t} - \nabla . [T_w(\nabla P_w + \rho_w gk)] = Q_w$$

Eqn. (3a)

$$\varphi \frac{\partial S_o}{\partial t} - \nabla . [T_o(\nabla P_o + \rho_o gk)], = Q_o$$

Eqn. (3b)

$$\nabla . \left(\frac{\rho_g}{B_g}T_g(\nabla P_g + \rho_g gk) + \frac{R_{so}\rho_g}{B_o}T_o(\nabla P_o + \rho_o gk)\right) - Q_g = -\frac{\partial}{\partial t}\left[\varphi\left(\frac{\rho_g}{B_g}S_g + \frac{R_{so}\rho_g}{B_o}S_o\right)\right]$$

Eqn. (3c)

where subscript $w$ = water, $o$ = oil, $g$ = gas. The system is closed by adding three additional equations.

$$P_{cwo} = P_o - P_w \; ; P_{cog} = P_g - P_o; S_w + S_o + S_g = 1$$

Eqn. (3d)

Gravity effects are considered by the terms $\rho_w gk$ and $\rho_o gk$. $\Omega \subset \mathbb{R}^n$ ($n = 2, 3$) is the modelling domain with boundary $\partial\Omega$, and $[0, t_f]$ is the time interval for which production data is available. $\varphi(x)$ = porosity, $T_o, T_w, T_g$ and $T$ the transmissibilities,

$$T_w = \frac{K(x)K_{rw}(S_w)}{\mu_w} \; ; T_o = \frac{K(x)K_{ro}(S_w)}{\mu_o} \; ; T_g = \frac{K(x)K_{rg}(S_o)}{\mu_g};$$

$$T = T_w + T_o + T_g$$

$K_{rw}(S_w)$ and $K_{ro}(S_w)$ = relative permeabilities, $\mu_w$, $\mu_o$, $\mu_g$ = viscosities of each phase. $Q_o$, $Q_w$, $Q_g$ and $Q = Q_o + Q_w + Q_g$ = oil, water flow, gas flow and the total flow, respectively. Eqn.3 (a-d) is solved with appropriate initial conditions, $S_w(x,0) = S_w^0(x)$; $P(x,0) = P_w^0(x)$ and a no-flux boundary condition, $\nabla p.v = 0$, on $\partial \Omega$. Gravity terms $\rho_w gk$ and $\rho_o gk$, as well as capillary pressure are ignored (such that $P_w = P_o = P_g = P$).

### 1.3 Surrogate forward modelling with a Fourier Neural operator (FNO) - Contribution

If the notation above is clear, we now discuss the *PINO* model used for solving the forward problem posed in Eqn.2(a & b) and Eqn. 3(a-d).

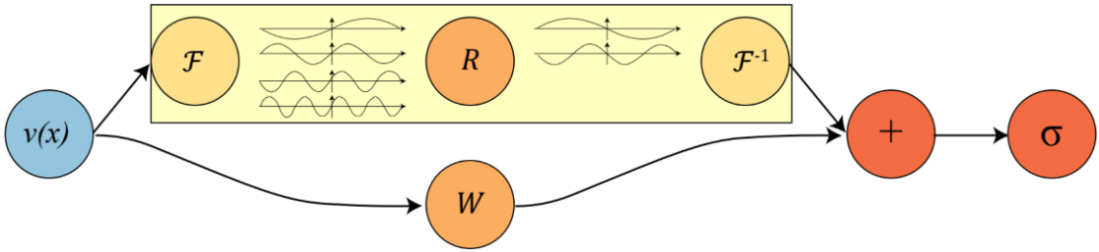

***Figure 1**: A standard Fourier neural operator (FNO) as adopted from [54]*

The Fourier neural operator (*FNO*) is effective in learning the mapping relationships between the infinite-dimensional function spaces that convert a traditional convolutional operation into a multiplication operation using the Fourier transform [54]. This causes the higher modes of the neural layer to be removed from the Fourier space, leaving only the lower modes in the Fourier layer. Hence, *FNO* can greatly improve the computational efficiency of the training process. The goal of *FNO* is to use a neural network $G^\dagger(a) = u$ to approximate the mapping of function $G : U \to Y, G \approx G^\dagger$, where Y and U are separable Banach spaces of function.

Figure 1 shows the full *FNO* architecture; input observation $v(x)$ is first lifted to a higher dimensional potential representation $e_0(x) = M(v(x))$ by fully connected shallow neural network $M$. Then, $e_0(x)$ is used as input for an iterative Fourier layer $e_0 \to e_1 \to e_2 \to \cdots \to e_N$. The concrete iterative process from state $e_n \to e_n + 1$ can be written as:

$$e_{n+1}(x) = \sigma\left(We_n(x) + \left(\wp(a;\vartheta)e_n(x)\right)(x)\right)$$

Eqn. (4)

where $W$ is a local liner transform; $\sigma$ is a non-linear activation function; $e_0$ and $e_1$ are the input and output of Fourier layer 1. Here, $\sigma$ is a non-linear activation function, $W$ is a linear transformation, and $\wp$ is an integral kernel transformation parameterized by a neural network. $\left(\wp(a;\vartheta)e_n(x)\right)(x)$ is the bias, that can be written as an integral:

$$\left(\wp(u;\vartheta)e_n(x)\right)(x) = \int \wp_\vartheta(x,y)e_n(y)dy$$

Eqn. (5)

To accelerate the integration process $\int \wp_\vartheta(x,y)e_n(y)dy$ Li et al. (2020) imposed a condition that $\wp_\vartheta(x,y) = \wp_\vartheta(x-y)$ and the new integral becomes,

$$(\wp(u; \vartheta)e_n(x))(x) = \int \wp_\vartheta(x,y)e_n(y)dy = F^{-1}(R_\vartheta \cdot (Fe_n))(x)$$

Eqn. (6)

where F denotes the Fourier transform, and $F^{-1}$ denotes the inverse Fourier transform. $R_\vartheta$ is a Fourier transform of periodic function $\wp$, $R_\vartheta = F(\wp\vartheta)$, and $\wp\vartheta$ is the kernel of the neural network. Following the N Fourier layer iteration, the last state output $e_N(x)$ is projected back to the original space using a fully connected neural network $S, y(x) = S(e_N(x))$.

In every block of the Fourier layer, *FNO* approximates highly non-linear functions mainly via combining the linear transform $W$ and global integral operator $R_\vartheta$ (Fourier transform). $R_\vartheta$ can be parameterized as a complex-valued tensor in the Fourier space from truncating high Fourier frequency modes $k$, and it can greatly reduce the number of trainable parameters and improve efficiency during the process of training.

### 1.3.1 Two phase flow physics informed surrogate modelling.

We are interested in predicting the pressure and water saturation given any input of the absolute permeability field $(K)$ and effective porosity $(\varphi)$. For completeness, the input and output tensors are indicated below, $p_{ini}, s_{ini}$ are the initial pressure and water saturation, $dt$, is the change in time steps. Two different *FNO* models [54] are trained simultaneously for the pressure and water saturation, with the same interacting loss function, this is to accommodate for the difference in magnitude between these quantities.

$$X = \{K, Q_w, Q, \varphi, dt, P_{ini}, S_{ini}\} \in \mathbb{R}^{B_0 \times 1 \times W \times H}(\text{for 2D case}), \mathbb{R}^{B_0 \times 1 \times D \times W \times H}(\text{for 3D case})$$
$$P = f_1(X; \theta_p), \quad S_w = f_2(X; \theta_s)$$

$K$ is the absolute permeability, $Q$ is the total flow, $Q_w$ is the water flow (injection rate), $dt$ is the time stepping, $P_{ini}$ is the initial pressure of the reservoir and $S_{ini}$ is the initial water saturation of the reservoir. $P$ is the pressure and $S_w$ is the water saturation. $f_1(X; \theta_p)$ is the *FNO* model for the pressure output and $f_2(X; \theta_s)$ is the *FNO* model for the saturation output.

The physics loss ansatz is then,

$$V(Q, p; T)_{pressure\ equation} = \frac{1}{n_s}(\| - \nabla \cdot [T\nabla P] - Q \|_2^2)$$

Eqn. (7a)

$$V(p, S_w; t, T_w)_{saturation\ equation} = \frac{1}{n_s} \left\| \left( \varphi \frac{\partial S_w}{\partial t} - \nabla \cdot [T_w(\nabla P)] \right) - Q_w \right\|_2^2$$

Eqn. (7b)

$$\phi_{cfd} = V(Q, p; T)_{pressure\ equation} + V(p, S_w; t, T_w)_{saturation\ equation}$$

Eqn.(8a)

where $n_s$ is the number of sample points. Together with the initial conditions and boundary conditions.

$$S_w(x, 0) = S_w^0(x) \ ; P(x, 0) = P(x) \ ; F_y = 0$$
$$\phi_{ic} = \frac{1}{n_s} \| S_w(x, 0) - S_w^0(x) \|_2^2 + \frac{1}{n_s} \| P(x, 0) - P(x) \|_2^2$$

Eqn. (8b)

$$\phi_{bc} = \frac{1}{n_s} \| F_y \|_2^2$$

Eqn.(8c)

$$\phi = \phi_{cfd} + \phi_{ic} + \phi_{bc}$$

Eqn. (9a)

$$\theta = [\theta_p, \theta_s]^T$$
$$\theta^{j+1} = \theta^j - \epsilon \nabla \phi_\theta^j$$

Eqn. (9b)

where j is the iteration step and $\epsilon > 0$ is the learning rate. The final $f_1(X; \theta_p)$, and $f_2(X; \theta_s)$ is the learned

*PINO* surrogate reservoir model. Also,
$$\nabla P = [P_x, P_y]$$

---

**Algorithm 1: Two-phase *PINO***

Input: $X_0 = \{K, Q_w, Q, \varphi, dt, P_{ini}, S_{ini}\} \in$
$\mathbb{R}^{B_0 \times 1 \times W \times H}$(for 2D case) or $\mathbb{R}^{B_0 \times 1 \times D \times W \times H}$(for 3D case),
$X_1 = \{K, Q_w, Q, \varphi, dt, P_{ini}, S_{ini}\} \in \mathbb{R}^{B_1 \times 1 \times W \times H}$(for 2D case) or $\mathbb{R}^{B_0 \times 1 \times D \times W \times H}$(for 3D case),
$Y_{1pt}$, --labelled pressure
$Y_{1st}$, -- labelled saturation
$f_1(:, \theta_p)$,
$f_2(:, \theta_s)$,
$T = 30$, -- Time
epoch, tol, $w_1, w_2, w_3, w_4, w_5, w_6 \in$

$j = 0$
while $(j \leq \text{epoch})$ or $(\phi \leq \text{tol})$ do
1. $Y_{0p} = f_1(X_0; \theta_p), Y_{0s} = f_2(X_0; \theta_s)$
2. $Y_{1p} = f_1(X_1; \theta_p), Y_{1s} = f_2(X_1; \theta_s)$,
3. Compute: $T_w = \frac{K(x)K_{rw}(Y_{0s})}{\mu_w}$ ; $T_o = \frac{K(x)K_{ro}(Y_{0s})}{\mu_o}$ ; $T = T_w + T_o$
4. Compute: $V(Y_{0p}, Y_{0s}; t, T_w)_{0s} = \frac{1}{n_s} \left\| \left( \varphi \frac{\partial Y_{0s}}{\partial t} - \nabla \cdot [T_w(\nabla Y_{0p})] \right) - X_0[Q_w] \right\|_2^2$
5. Compute: $V(X_0[Q], Y_{0p}; T)_{0p} = \frac{1}{n_s} \left( \left\| -\nabla \cdot [T\nabla Y_{0p}] - X_0[Q] \right\|_2^2 \right)$
6. Compute: $T_w = \frac{K(x)K_{rw}(Y_{1s})}{\mu_{water}}$ ; $T_o = \frac{K(x)K_{ro}(Y_{1s})}{\mu_{oil}}$ ; $T = T_w + T_o$
7. Compute: $V(Y_{1p}, Y_{1s}; t, T_w)_{1s} = \frac{1}{n_s} \left\| \left( \varphi \frac{\partial Y_{1s}}{\partial t} - \nabla \cdot [T_w(\nabla Y_{1p})] \right) - X_1[Q_w] \right\|_2^2$
8. Compute: $V(X_1[Q], Y_{1p}; T)_{1p} = \frac{1}{n_s} \left( \left\| -\nabla \cdot [T\nabla Y_{1p}] - X_1[Q] \right\|_2^2 \right)$
9. $\phi_p = \left\| Y_{1pt,} - f_1(X_1; \theta_p) \right\|_2^2$
10. $\phi_s = \left\| Y_{1st,} - f_2(X_1; \theta_s) \right\|_2^2$
11. $\phi = w_1\phi_p + w_2\phi_s + w_3V(Y_{0p}, Y_{0s}; t, T_w)_{0s} + w_4V(X_0[Q], Y_{0p}; T)_{0p} + w_5V(Y_{1p}, Y_{1s}; t, T_w)_{1s} + w_6V(X_1[Q], Y_{1p}; T)_{1p}$
12. Update models:
$$\theta = [\theta_p, \theta_s]^T$$
$$\theta^{j+1} = \theta^j - \epsilon \nabla \phi_\theta^j$$
   $j \leftarrow j + 1$
Output: $f_1(:, \theta_p), f_2(:, \theta_s)$,

---

$w_1, w_2, w_3, w_4, w_5, w_6$ are the weights associated to the loss functions associated to the 6 terms. $X_0 = \{K, Q_w, Q, \varphi, dt, P_{ini}, S_{ini}\} \in \mathbb{R}^{B_0 \times 1 \times W \times H}$ are the dictionary inputs without running the numerical reservoir simulator. $X_1 = \{K, Q_w, Q, \varphi, dt, P_{ini}, S_{ini}\} \in \mathbb{R}^{B_1 \times 1 \times W \times H}$ are the dictionary inputs from running the reservoir simulator. $epoch, tol$ are the number or epochs and the tolerance level for the loss function. $f(:, \theta_p), f(:, \theta_s)$ are the FNO models for the pressure equation and water saturation equation respectively.

### 1.3.2 Three phase flow physics informed surrogate modelling.

We are interested in predicting the pressure, water saturation and gas saturation fields given any input of the absolute permeability field ($K$) and effective porosity ($\varphi$). For completeness, the input and output tensors are indicated below, $p_{ini}, s_{ini}$ are the initial pressure and water saturation, $dt$, is the change in time steps. three different *FNO* models [54] are trained simultaneously for the pressure, water saturation and gas saturation, with the same interacting loss function, this is to accommodate for the difference in magnitude between these quantities.
$$X = \{K, FTM, \varphi, P_{ini}, S_{ini}\} \in \mathbb{R}^{B_0 \times 1 \times D \times W \times H}$$

$$P = f_1(X; \theta_p), \quad S_w = f_2(X; \theta_s), \quad S_g = f_3(X; \theta_g),$$

$K$ is the absolute permeability, $Q$ is the total flow, $Q_w$ is the water flow (injection rate), $dt$ is the time stepping, $P_{ini}$ is the initial pressure of the reservoir and $S_{ini}$ is the initial water saturation of the reservoir. $P$ is the pressure and $S_w$ is the water saturation. $f_1(X; \theta_p)$ is the FNO model for the pressure output and $f_2(X; \theta_s)$ is the FNO model for the water saturation output and $f_3(X; \theta_g)$ is the FNO model for the gas saturation output.

The physics loss is then,

$$V(Q, p; T)_{pressure\ equation} = \frac{1}{n_s}(\|- \nabla \cdot [T\nabla P] - Q\|_2^2)$$

Eqn. (10a)

$$V(p, S_w; t, T_w)_{water\ equation} = \frac{1}{n_s}\left\|\left(\varphi \frac{\partial S_w}{\partial t} - \nabla \cdot [T_w(\nabla P)]\right) - Q_w\right\|_2^2$$

Eqn. (10b)

$$V(p, S_g; t, T_g)_{gas\ equation} = \frac{1}{n_s}\left\|\nabla \cdot \left(\frac{\rho_g}{B_g}T_g(\nabla P) + \frac{R_{so}\rho_g}{B_o}T_o(\nabla P)\right) - Q_g + \frac{\partial}{\partial t}\left[\varphi\left(\frac{\rho_g}{B_g}S_g + \frac{R_{so}\rho_g}{B_o}S_o\right)\right]\right\|_2^2$$

Eqn.(10c)

$$\phi_{cfd} = V(Q, p; T)_{pressure\ equation} + V(p, S_w; t, T_w)_{water\ equation} + V(p, S_g; t, T_g)_{gas\ equation}$$

Eqn. (11)

Together with the initial conditions and boundary conditions.

$$S_w(x, 0) = S_w^0(x) \ ; \ P(x, 0) = P(x); S_g(x, 0) = S_g^0(x); F_y = 0$$

$$\phi_{ic} = \frac{1}{n_s}\|S_w(x, 0) - S_w^0(x)\|_2^2 + \frac{1}{n_s}\|P(x, 0) - P(x)\|_2^2 + \frac{1}{n_s}\|S_g(x, 0) - S_g^0(x)\|_2^2$$

Eqn. (12a)

$$\phi_{bc} = \frac{1}{n_s}\|F_y\|_2^2$$

Eqn.(12b)

$$\phi = \phi_{cfd} + \phi_{ic} + \phi_{bc} + \phi_{data}$$

Eqn. (13)

$$\theta = [\theta_p, \theta_s, \theta_g]^T$$
$$\theta^{j+1} = \theta^j - \epsilon \nabla \phi_\theta^j$$

In algorithm 2, $w_1 \ldots, w_8$ are the weights associated to the loss functions associated to the 8 terms. $X_0 = \{K, \text{FTM}, \varphi, P_{ini}, S_{ini}\}$ are the dictionary inputs without running the numerical reservoir simulator. $X_1 = \{K, \text{FTM}, \varphi, P_{ini}, S_{ini}\}$ are the dictionary inputs from running the reservoir simulator. $epoch, tol$ are the number or epochs and the tolerance level for the loss function. $f_1(:, \theta_p), f_2(:, \theta_s), f_3(:, \theta_g)$ are the FNO models for the pressure, water, and gas saturation equation respectively.

### 1.4 Inverse Problem

In this section, we lay out the problem setting for the inverse problem by primarily following the mathematical setup in [56]. We follow the notation of section 1.2 with $G: U \rightarrow Y$ as the forward operator. We aim to solve an inverse problem of finding unknown $u \in U$ given observations y of the form.

$$y = G(u) + \eta$$

Eqn. (14)

Algorithm 2: **Three-phase *PINO***

Input: $X_0 = \{K, FTM, \varphi, P_{ini}, S_{ini}\} \in \mathbb{R}^{B_0 \times 1 \times D \times W \times H}, X_{N0} = \{Q_w, Q_g, Q, dt\} \in \mathbb{R}^{B_0 \times T \times D \times W \times H}$
$X_1 = \{K, FTM, \varphi, P_{ini}, S_{ini}\} \in \mathbb{R}^{B_1 \times 1 \times D \times W \times H}, , X_{N1} = \{Q_w, Q_g, Q, dt\} \in \mathbb{R}^{B_1 \times T \times D \times W \times H}$
$Y_{1pt,}$ -- labelled pressure
$Y_{1st,}$ -- labelled water saturation
$Y_{1gt,}$ -- labelled gas saturation
$f_1(:, \theta_p),$
$f_2(:, \theta_s),$
$f_3(:, \theta_g),$
T -- Time
epoch, tol, $w_1, w_2, w_3, w_4, w_5, w_6 \in$
$j = 0$
while $(j \leq \text{epoch})$ or $(\phi \leq \text{tol})$ do
   1. $Y_{0p} = f_1(X_0; \theta_p), Y_{0s} = f_2(X_0; \theta_s), Y_{0g} = f_3(X_0; \theta_g)$
   2. $Y_{1p} = f_1(X_1; \theta_p), Y_{1s} = f_2(X_1; \theta_s), Y_{1g} = f_3(X_1; \theta_g),$
   3. Compute: $T_w = \frac{K(x) K_{rw}(Y_{0s})}{\mu_w}; T_o = \frac{K(x) K_{ro}(Y_{0s})}{\mu_o}; T_g = \frac{K(x) K_{rg}(1 - (Y_{0s} + Y_{0g}))}{\mu_g}; T = T_w + T_o + T_g$
   4. Compute: $V(Y_{0p}, Y_{0s}; t, T_w)_{0s} = \frac{1}{n_s} \left\| \left( \varphi \frac{\partial Y_{0s}}{\partial t} - \nabla \cdot [T_w(\nabla Y_{0p})] \right) - X_{N0}[Q_w] \right\|_2^2$
   5. Compute: $V(Y_{0p}, Y_{0s}, Y_{0g}; t, T_o, T_g)_{0g} = \frac{1}{n_s} \left\| \nabla \cdot \left( \frac{\rho_g}{B_g} T_g(\nabla Y_{0p}) + \frac{R_{so} \rho_g}{B_o} T_o(\nabla Y_{0p}) \right) - X_{N0}[Q_g] + \frac{\partial}{\partial t} \left[ \varphi \left( \frac{\rho_g}{B_g} Y_{0g} + \frac{R_{so} \rho_g}{B_o} (1 - (Y_{0s} + Y_{0g})) \right) \right] \right\|_2^2$
   6. Compute: $V(X_0[Q], Y_{0p}; T)_{0p} = \frac{1}{n_s} \left( \| -\nabla \cdot [T \nabla Y_{0p}] - X_0[Q] \|_2^2 \right)$
   7. Compute: $T_w = \frac{K(x) K_{rw}(Y_{1s})}{\mu_w}; T_o = \frac{K(x) K_{ro}(Y_{1s})}{\mu_o}; T_g = \frac{K(x) K_{rg}(1 - (Y_{1s} + Y_{1g}))}{\mu_g}; T = T_w + T_o + T_g$
   8. Compute: $V(Y_{1p}, Y_{1s}; t, T_w)_{1s} = \frac{1}{n_s} \left\| \left( \varphi \frac{\partial Y_{1s}}{\partial t} - \nabla \cdot [T_w(\nabla Y_{1p})] \right) - X_1[Q_w] \right\|_2^2$
   9. Compute: $V(Y_{1p}, Y_{1s}, Y_{1g}; t, T_o, T_g)_{1g} = \frac{1}{n_s} \left\| \nabla \cdot \left( \frac{\rho_g}{B_g} T_g(\nabla Y_{1p}) + \frac{R_{so} \rho_g}{B_o} T_o(\nabla Y_{1p}) \right) - X_{N1}[Q_g] + \frac{\partial}{\partial t} \left[ \varphi \left( \frac{\rho_g}{B_g} Y_{1g} + \frac{R_{so} \rho_g}{B_o} (1 - (Y_{1s} + Y_{1g})) \right) \right] \right\|_2^2$
   10. Compute: $V(X_1[Q], Y_{1p}; T)_{1p} = \frac{1}{n_s} \left( \| -\nabla \cdot [T \nabla Y_{1p}] - X_1[Q] \|_2^2 \right)$
   11. $\phi_p = \| Y_{1pt,} - f_1(X_1; \theta_p) \|_2^2$
   12. $\phi_s = \| Y_{1st,} - f_2(X_1; \theta_s) \|_2^2$
   13. $\phi_g = \| Y_{1gt,} - f_3(X_1; \theta_g) \|_2^2$
   14. $\phi = w_1 \phi_p + w_2 \phi_s + w_3 \phi_g + w_4 V(Y_{0p}, Y_{0s}; t, T_w)_{0s} + w_5 V(X_0[Q], Y_{0p}; T)_{0p} + w_6 V(Y_{1p}, Y_{1s}; t, T_w)_{1s} + w_7 V(X_1[Q], Y_{1p}; T)_{1p} + w_8 V(Y_{0p}, Y_{0s}, Y_{0g}; t, T_o, T_g)_{1g}$
   15. Update models:
$$\theta = [\theta_p, \theta_s, \theta_g]^T$$
$$\theta^{j+1} = \theta^j - \epsilon \nabla \phi_\theta^j$$
   $j \leftarrow j + 1$
Output: $f_1(:, \theta_p), f_2(:, \theta_s), f_3(:, \theta_g),$

where η is assumed a Gaussian noise with known covariance Γ. To solve the inverse problem instead of a filtering problem with Ensemble Kalman methods, we construct an artificial dynamic based on state augmentation by constructing the space $Z = Y \times U$, and the mapping $\Xi: Z \to Z$, where.

$$\Xi(z) = \begin{pmatrix} u \\ G(u) \end{pmatrix} \text{ for } z = \begin{pmatrix} u \\ y \end{pmatrix} \in Z.$$

Eqn. (15)

Thus, we have the following artificial dynamics $z_{n+1} = \Xi(z_n)$. The observation y includes all the historical observed data in a fixed time window. We emphasize that the artificial dynamics $\Xi$ is the iterative scheme of the ensemble Kalman inversion process, which is different from the actual time dependent dynamics of the oil reservoir model.

We denote the observation operator by $H: Z \to Y$ which is defined as $H = (0, I)$. We assume the following relationship between the observation and the dynamics,

$$y_{n+1} = H z_{n+1} + \eta_{n+1}$$

Eqn. (16),

where $\{\eta_n\}_{n \in Z^+}$ is an i.i.d. sequence with $\eta_n \sim N(0, \Gamma)$. We denote the prior distribution by $\mu_0$. It is of common practice to consider a log normal prior on the permeability field. In such case, the Gaussian random permeability field can be represented by $K = u_0 + \exp\left(\sum_{j \leq 1} u_j \sqrt{\{\lambda_j\}} \theta_j\right)$ where $\lambda_j$ and $\theta_j$ are the KL expansion basis and $u_0, u_j$ are real coefficients. Given a log normal Gaussian prior $\mu_0$, then we can sample K with $u_j \sim N(0, 1)$. We denote by $\gamma_1$ the standard Gaussian measure in $R^1$, then we have the probability space $(R^N, B(R)^N, \gamma)$, where B is the Borel σ-algebra and $\gamma = \otimes_{i=1}^{\infty} \gamma_1$. The wellposedness for elliptic as well as parabolic type problems with log normal Gaussian prior has been investigated in [35,57]. The well-posedness for flow equation is also investigated in [58] in two dimensions with periodic boundary condition. However, rigorous proofs for the black oil reservoir models listed in Section 1.2 are not available to the author's best knowledge. Despite the elliptic-parabolic like coupled system, it is complicated by the facts that the coefficient in the elliptic equation (e.g. the viscosity in the pressure equation) is dependent on the saturation [59]. In addition, non-Gaussian priors have been widely used and validated within the engineering community. In section 3, we investigated our method with numerical experiments for history matching inverse problem with both Gaussian prior and exotic prior. Rigorous analysis of ensemble Kalman methods for inverse problem can be found in [56] with numerical examples on Darcy model of groundwater flow. Here we adopted a variation of vanilla ensemble Kalman inversion method called αREKI described in Algorithm 3.

Typically, in the application of ensemble Kalman methods for history matching problems of oil reservoir modelling, the construction of the Kalman gain matrix is often computationally cheaper relative to the computational cost of the forward operator G, especially for high fidelity numerical models. In this paper, we propose the use of AI surrogate model $G^\dagger$ discussed in section 1.3 to replace the numerical approximation of G in Algorithm 3.

The ensemble update step can be improved by the application of techniques known as covariance localization. It is a technique used in reservoir history matching to filter divergence elimination and spurious correlations, which are found in both standard ensemble-based methods. It selectively assimilates data during the update step by adjusting the correlation between the data measured at a location A and the model parameters or data measured at a further location B to be zero, therefore eliminating long-distance non-zero spurious correlations and increasing the degrees of freedom available for data assimilation. Various localization methods are available in the literature, and each scheme distinguishes themselves by the choice of the localizing function. In this paper, we adopt the fifth order piecewise rational function by Gaspari and Cohn as shown below,

Algorithm 3: *a*REKI

Input: $\{u_0^{(j)}\}_{j=1}^{J}$ -- prior
y -- measurements
$\Gamma$ --measurement errors covariance
iter -- iteration

$s_n = 0$
$n = 0$
while $(s_n < 1)$ or $(n \leq \text{iter})$ do
  1. Compute: $G(u_n^{(j)}), j \in \{1,\ldots J\}$
  2. Compute: $\Phi_n \equiv \left\{\frac{1}{2}\left\|\Gamma^{-\frac{1}{2}}(y - G(u))\right\|_2^2\right\}_{j=1}^{J}$
  3. Compute: $\overline{\Phi_n}, \sigma_{\Phi_n}^2$ -- mean & variance
  4. Compute: $\alpha_n = \dfrac{1}{\min\left\{\max\left\{\frac{L_{(y)}}{2\overline{\Phi_n}}, \sqrt{\frac{L_{(y)}}{2\sigma_{\Phi_n}^2}}\right\}, 1 - s_n\right\}}$
  5. Compute: $s_n = s_n + \frac{1}{\alpha_n}$
  6. Compute: $C_n^{GG} = \frac{1}{j-1}\sum_{j=1}^{J}\left(G(u_n^{(j)}) - \overline{G(u_n^{(j)})}\right) \otimes \left(G(u_n^{(j)}) - \overline{G(u_n^{(j)})}\right), j \in \{1,\ldots J\}$
  7. Compute: $C_n^{uG} = \frac{1}{j-1}\sum_{j=1}^{J}\left((u_n^{(j)}) - \overline{(u_n^{(j)})}\right) \otimes \left(G(u_n^{(j)}) - \overline{G(u_n^{(j)})}\right), j \in \{1,\ldots J\}$
  8. Update ensemble :
    $u_{n+1}^{(j)} = u_n^{(j)} + C_n^{uG}(C_n^{GG} + \alpha_n\Gamma)^{-1}\left(y + \sqrt{\alpha_n}\xi_n - G(u_n^{(j)})\right), j \in \{1,\ldots J\}$

$n \leftarrow n + 1$
Output: $u^{(j)}$

$$\rho = \begin{cases} -\frac{1}{4}\left(\frac{|z|}{c}\right)^5 + \frac{1}{2}\left(\frac{|z|}{c}\right)^4 + \frac{5}{8}\left(\frac{|z|}{c}\right)^3 - \frac{5}{3}\left(\frac{|z|}{c}\right)^2 + 1, & 0 \leq |z| \leq c, \\ \frac{1}{12}\left(\frac{|z|}{c}\right)^5 - \frac{1}{2}\left(\frac{|z|}{c}\right)^4 + \frac{5}{8}\left(\frac{|z|}{c}\right)^3 + \frac{5}{3}\left(\frac{|z|}{c}\right)^2 - 5\left(\frac{|z|}{c}\right) + 4 - \frac{2}{3}\left(\frac{|z|}{c}\right), & c \leq |z| \leq 2c, \\ 0, & 2c \leq |z|. \end{cases}$$

Eqn. (17)

$$u_{n+1}^{(j)} = u_n^{(j)} + \rho \circ C_n^{uG}\left(C_n^{GG} + \alpha_n\Gamma\right)^{-1}\left(y + \sqrt{\alpha_n}\xi_n - G(u_n^{(j)})\right), j \in \{1,\ldots J\}$$

Eqn. (18)

The localization function can be introduced to the ensemble update step by Schur product, which is an element-by-element multiplication [15]. It is introduced to modify the Kalman Gain during update step for *a*REKI.

## 1.5 Mixture of experts.

In general, the problem of designing machine learning based models in a supervised context is the following. Assume $\{(x_i, y_i)\}_{i=1}^{N}$ where $x_i \in \mathbb{R}^K$ and $y_i \in \mathbb{R}^M$ for regression or $y_i \in \{0,1,\ldots J\}^M$ for classification. Postulating amongst a family of ansatz $f(.;\theta)$, parametrized by $\theta \in \mathbb{R}^P$, we can find a $\theta^*$ such that for all $i = 1,\ldots N, y_i = f(x_i; \theta^*)$. Then for all $x'$ in the set $\{x_i\}_{i=1}^{N}$, $y' \approx f(x'; \theta^*)$. Where $y'$ is the true label of $x'$

For a set of labelled data $D = \{(x_i, y_i)\}_{i=1}^{N}$, where $(x_i, y_i) \in \chi \times \gamma$ assumed to be input and output of a model shown in Eqn. 19.

$$y_i \approx f(x_i)$$

Eqn. 19

$f: \chi \rightarrow \gamma$ is irregular and has sharp features, very non-linear and has noticeable discontinuities,
The output space is taken as $\gamma = \mathbb{R}$ and $\chi = \mathbb{R}^d$

### 1.5.1 Cluster

In this stage, we seek to cluster the training input and output pairs.
$\lambda: \chi \times \gamma \rightarrow \mathcal{L} := \{1, \ldots, L\}$ where the label function minimizes,

$$\Phi_{clust}(\lambda) = \sum_{l=1}^{L} \sum_{i \in S_l} \ell(x_i, y_i)$$

Eqn. 20

$$S_l = \{(x_i, y_i); \lambda(x_i, y_i) = l\}$$

Eqn. 21

$\ell_l$ loss function associated to cluster $l$

$$z_i = (x_i, y_i), \ell_l = |z_i - \mu_l|^2$$

Eqn. 22

$\mu_l = \frac{1}{|S_l|} \sum_{i \in S_l} z_i$ where $|.|$ denotes the Euclidean norm

### 1.5.2 Classify

$l_i = \lambda(x_i, y_i)$ is an expanded training set

$$\{(x_i, y_i, l_i)\}_{i=1}^{N}$$

Eqn. (23a)

$$f_c: \chi \rightarrow \mathcal{L}$$

Eqn. (23b)

$x \in \chi$ provides an estimate $f_c : x \mapsto f(x) \in \mathcal{L}$ such that $f_c(x_i) = l_i$ for most of the data. Crucial for the ultimate fidelity of the prediction. $\{y_i\}$ is ignored at this phase. The classification function minimizes,

$$\Phi_{clust}(f_c) = \sum_{i=1}^{N} \phi_c(l_i, f_c(x_i))$$

Eqn. (23c)

$\phi_c : \mathcal{L} \times \mathcal{L} \rightarrow \mathbb{R}_+$ is small if $f_c(x_i) = l_i$ for example we can choose $f_c(x) = argmax_{l \in \mathcal{L}} g_l(x)$ where $g_l(x) > 0, \sum_{l=1}^{L} g_l(x)$ is a soft classifier and $\phi_c(l_i, f_c(x_i)) = -\log(g_l(x))$ is a cross-entropic loss.

### 1.5.3 Regress

$$f_r : \chi \times \mathcal{L} \rightarrow \gamma$$

Eqn. (24a)

For each $(x, l) \in \chi \times \mathcal{L}$ must provide an estimate $f_r : (x, l) \mapsto f_r(x, l) \in \gamma$ such that $f_r(x, f_c(x)) \approx y$ for both the training and test data. If successful a good reconstruction for

$$f: \chi \rightarrow \gamma$$

Eqn. (24b)

Where $f(.) = f_r(., f_c(.))$ the regression function can be found by minimizing

$$\Phi_r(f_r) = \sum_{i=1}^{N} \phi_r\left(y_i, f_r(x_i, f_c(x_i))\right)$$

Eqn. (24c)

Where $\phi_r : \gamma \times \gamma \rightarrow \mathbb{R}_+$ minimized when $f_r(x_i, f_c(x_i)) = y_i$ in this case can be chosen as $\phi_r\left(y, f_r(x, f_c(x))\right) = |y - f_r(x, f_c(x))|^2$. Data can be partitioned into $C_l = \{i; f_c(x_i) = l\}$ for $l = 1, \ldots L$ and then perform $L$ separate regressions done in parallel.

$$\Phi_r^l(f_r(., l) = \sum_{i \in C_l} \phi_r(y_i, f_r(x_i, l))$$

Eqn. (24d)

### 1.5.4 Scaling of the Data

$x = (x^1, \ldots x^d) \in \mathbb{R}^d$ and $y \in \mathbb{R}$ hence, have $|(x,y) - (x',y')|^2 = (y - y')^2 + |x - x'|^2$

For $j = 1, \ldots d$,

$$\tilde{x}^j = \left(x^j - min_{i \in \{1,..N\}} x_i^j\right) / \left(max_{i \in \{1,..N\}} x_i^j - min_{i \in \{1,..N\}} x_i^j\right)$$

Eqn. (25a)

$$\tilde{y} = C\left(y - min_{i \in \{1,..N\}} y_i\right) / \left(max_{i \in \{1,..N\}} y_i - min_{i \in \{1,..N\}} y_i\right)$$

Eqn. (25b)

For $C > 1$ $C = 10d$ where $d = \dim(x)$ For regression $C = 1$

### 1.5.5 Bayesian Formulation

A critique of this method is that it re-uses the data in each phase. A Bayesian postulation handles this limitation elegantly. Recall,

$$\boldsymbol{D} = \{(x_i, y_i)\}_{i=1}^N$$

Eqn. (26a)

Assume parametric models for the classifier $g_l(.;\theta_c) = g_l(.;\theta_c^l)$ and the regressor. $f_r(.,l;\theta_r^l)$ for $l = 1, \ldots L$ where $\theta_c = (\theta_c^1, \ldots, \theta_c^L)$ and $\theta_r = (\theta_r^1, \ldots, \theta_r^L)$ and let $\theta = (\theta_c, \theta_r)$, the posterior density has the form,

$$\pi(\theta, l|D) \propto \prod_{i=1}^N \pi(y_i|x_i, \theta_r, l) \, \pi(l|x_i, \theta_c) \pi(\theta_r) \pi(\theta_c)$$

Eqn. (26b)

$$\pi(y_i|x_i, \theta_r, l) \propto \exp\left(-\frac{1}{2}|y_i - f_r(x_i, l; \theta_r^l)|^2\right)$$

Eqn. (26c)

And

$$\pi(l|x_i, \theta_c) = g_l(x_i; \theta_c)$$

Eqn. (26d)

$$g_l(x_i; \theta_c) = \frac{\exp\left(h_l(x; \theta_c^l)\right)}{\sum_{l=1}^L \exp\left(h_l(x; \theta_c^l)\right)}$$

Eqn. (26e)

$h_l(x; \theta_c^l)$ are some standard parametric regressors. We note that the 1-pass CCR algorithm can be used as parametrisation method to maintain discontinuous functions or patters in the unknown parameter (permeability & porosity ) field.

### 1.6 NVIDIA Modulus

NVIDIA Modulus is an open-source deep-learning framework for building, training, and fine-tuning deep learning models using state-of-the-art physics ML methods [60]. This framework is part of NVIDIA's suite of AI tools and is tailored for domain scientists, engineers, and AI researchers working on complex simulations across various domains such as fluid dynamics, atmospheric sciences, structural mechanics, electromagnetics, acoustics, and related domains. The emphasis on providing components and models that introduce inductive bias into the neural networks (e.g. FNOs, GNNs, etc.) and training (e.g. physics-based constraints) allows for more accurate and generalizable trained models for physics-ML applications, especially in scenarios where the data might be scarce or expensive to obtain.

The Modulus framework is fundamentally composed of Modulus Core (generally referred to as Modulus) and Modulus-Sym. Modulus-Sym is built on top of Modulus which is a generalized toolkit for physics-ML model development and training. Modulus provides core utilities to train AI models in the physics-ML domain with optimized implementations of various network architectures specifically suited for physics-based applications (MLPs, operator networks like FNOs, DeepONet, graph networks), efficient data pipes for typical physics-ML datasets (structured and unstructured grids, point-clouds, etc.) and utilities to distribute and scale the training across multiple GPUs and nodes. Modulus-Sym provides utilities to introduce physics-based constraints into the model training and has utilities to handle various geometries (via built-in geometry module or STLs), introduce PDE constraints, and compute gradients efficiently through an optimized gradient backend. It also provides abstractions such as pre-defined training loops that enable domain scientists to experiment with AI model training quickly without diving into the depths of specific implementations of DL training. Modulus-Sym is designed to be highly flexible, supporting various types of partial differential equations (PDEs) (built-in PDE definitions as well as easy ways to introduce your own PDEs symbolically) and constraint definitions typically found in physics-based problems like boundary and equation residual conditions.

Modulus is compatible with NVIDIA's broader AI and HPC (high-performance computing) ecosystems. It is built on top of PyTorch and is interoperable with PyTorch. Modulus' architecture allows for easy customization and has several integrations with other NVIDIA technologies like CUDA, Warp, PySDF, DALI, NVFuser, and Omniverse, to facilitate efficient computation, scaling, performance optimizations, and real-time visualization leveraging NVIDIA GPUs.

**How It Works**
Modulus provides several tools required to set up an AI model training. Fundamentally, the goal of training an AI model using Modulus requires translating the problem of solving PDEs into a machine-learning task. Under the abstracted workflow of Modulus-Sym, the user defines the problem in terms of PDEs, boundary conditions, and any available data. The framework then employs neural networks to approximate the solution to these PDEs. During training, the loss function is designed to minimize not just the error between the predicted and actual data points (if available) but also the discrepancy from satisfying the physical laws as described by the PDEs and boundary/initial conditions. This dual emphasis on data fidelity and physical consistency is what makes PIML models powerful.

**Key Features**
- **Solver Independence**: Unlike traditional numerical solvers that are often specialized for specific types of PDEs or require grid-based discretization, Modulus is more flexible and capable of handling a wide range of problems without the need for meshing the domain.
- **GPU Acceleration**: Leveraging NVIDIA GPUs, Modulus enables the training of AI models and surrogates that can perform complex calculations and model training significantly faster than CPU-based approaches, enabling faster iteration times.
- **High-Level API**: Modulus-Sym provides a high-level API that abstracts away many of the complexities involved in setting up PIML models, making it accessible to users who may not be experts in machine learning or computational physics.

**Use Cases**
Practical applications of NVIDIA Modulus span multiple industries and research areas. With the flexibility to train models using data and physics constraints, NVIDIA Modulus has enabled several applications across multiple domains. For example, in the field of atmospheric sciences and fluid dynamics, AI models developed for global weather prediction, super-resolution of flow fields, industrial digital twins for fluid flow in and past complex geometries and shapes have showcased remarkable savings in computational cost while maintaining good accuracy compared to traditional methods. In energy, applications include modelling subsurface flows for oil and gas exploration and optimizing renewable energy systems.

NVIDIA Modulus represents a significant advancement in the field of computational science, marrying the predictive power of machine learning with the foundational principles of physics. Its ability to integrate physical laws into the learning process opens new possibilities for solving complex scientific problems, making it a valuable tool for researchers and engineers looking to push the boundaries of simulation and modelling.

## 2. Methodology

In the workflow in Fig. 2, a suitable training-image (TI) is used to generate an initial prior density of permeability fields using the MPSlib software library [18]. This prior density is then used to generate a non-Gaussian prior (VCAE/CCR). A selected number of realisations is then used for the inverse problem workload. First, we forward the ensemble members using the prior developed *PINO* model to get predicted data. Then we parametrise these permeability fields with the CCR/VCAE, next we correct these quantities with *a*REKI and we recover the original parameters to repeat the process until a suitable stopping criterion is met.

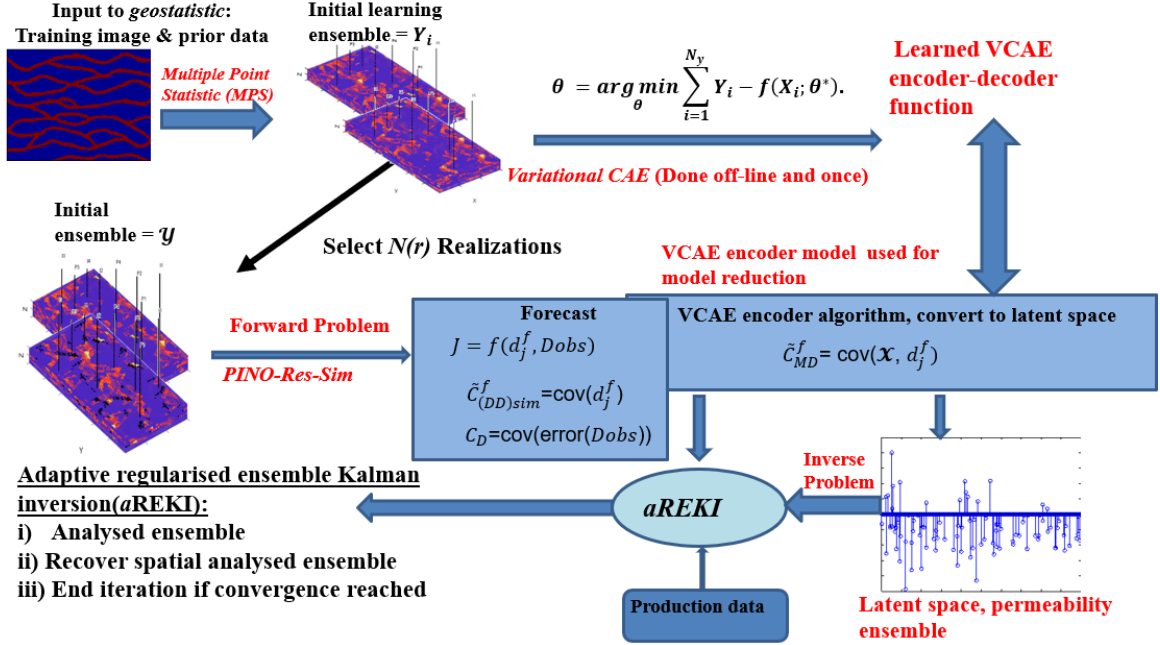

*Figure 2: Cartoon showing the overall developed PINO-aREKI workflow.*

In this methodology, we model the combined dynamic property estimation surrogate shown in algorithm (1 & 2) together with the peaceman analytical well model shown in Eqn. (27) via the CCR approach in section 1.5 to yield Eqn. (28)

$$Q = \frac{2\pi K k_r h \left( P_{avg} - P_{wf} \right)}{\mu B \ln\left(\frac{r_e}{r_w}\right) + s}$$

Eqn. (27)

Where:
- $Q$ is the flow rate.
- $Kk_r$ is the absolute permeability of the reservoir at the voxel location of the well.
- $k_r$ is the relative permeability which is a function of water and gas saturation.
- $h$ is the thickness of the reservoir.
- $P_{avg}$ is the average reservoir pressure.
- $P_{wf}$ is the bottom hole flowing pressure.
- $\mu$ is the viscosity of the fluid.
- $B$ is the formation volume factor.
- $Rr_e$ is the effective drainage radius.
- $r_w$ is the radius of the wellbore.
- $s$ is the skin factor.

This gives a novel implementation where we couple the CCR algorithm and the *PINO* approach to give a surrogate reservoir simulator called *PINO-Res-Sim*

$$f_{ccr}(f_1(:,\theta_p), f_2(:,\theta_s), f_3(:,\theta_g); \theta_{ccr})$$

Eqn. (28)

The detailed implementation is shown in Algorithm. 4 and it replaces the expensive numerical forward solver during the history matching loop.

---

Algorithm 4: ***PINO – Res – Sim*** Reservoir simulator

Input
$X_0 = \{K, FTM, \varphi, P_{ini}, S_{ini}\} \in \mathbb{R}^{B_1 \times 1 \times D \times W \times H}$
$f_1(:,\theta_p) - PINO$ pressure surrogate,
$f_2(:,\theta_s) - PINO$ Water saturation surrogate,
$f_3(:,\theta_g) - PINO$ Gas saturation surrogate,
$f_{ccr}(:,\theta_{ccr}) - $CCR Peaceman surrogate (section 1.4, CCR algorithm),
$N_{pr}$ – Number of producers
$N_x, N_y, N_z$ – size of reservoir computational domain
T – Time Index
$i_{pef}, j_{pef}, k_{pef}$ locations/completions

Begin
1. Compute: $Y_{1p} = f_1(X_0; \theta_p), Y_{1s} = f_2(X_0; \theta_s), Y_{1g} = f_3(X_0; \theta_g)$

2. Construct: $X_1 = \begin{bmatrix} \frac{1}{N_z}\sum_{k_{pef}=1}^{N_z} K_{i_{perf_n} j_{perf_n} k_{pef}}, \\ \frac{1}{N_{pr}}\sum_{n=1}^{N_{pr}} \frac{1}{N_z}\sum_{k_{pef}=1}^{N_z} f_1(X_1,\theta_p)_{i_{perf_n} j_{perf_n} k_{pef}}, \\ \frac{1}{N_z}\sum_{k_{pef}=1}^{N_z} f_2(X_1,\theta_s)_{i_{perf_n} j_{perf_n} k_{pef}}, \\ \frac{1}{N_z}\sum_{k_{pef}=1}^{N_z} f_3(X_1,\theta_g)_{i_{perf_n} j_{perf_n} k_{pef}}, \\ 1 - \left(\begin{array}{c}\frac{1}{N_z}\sum_{k_{pef}=1}^{N_z}\frac{1}{N_z}\sum_{k_{pef}=1}^{N_z} f_2(X_1,\theta_s)_{i_{perf_n} j_{perf_n} k_{pef}} \\ + \frac{1}{N_z}\sum_{k_{pef}=1}^{N_z} f_3(X_1,\theta_g)_{i_{perf_n} j_{perf_n} k_{pef}}\end{array}\right), \\ T \\ \forall n = 1,2 \ldots N_{pr} \end{bmatrix} \in \mathbb{R}^{B_1 \times T \times [4N_{pr}+2]}$

3. Compute: $Y_{out} = f_{ccr}(X_1; \theta_{ccr}) \in \mathbb{R}^{B_1 \times T \times 3N_{pr}}$ – wopr, wwpr, wgpr

Output: $Y_{out}$

---

## 3. Numerical Experiments
The code for these numerical experiments is staged in the links below,
- Detailed installation for NVIDIA Modulus-Sym can be found at:
  https://github.com/NVIDIA/modulus-sym/tree/main
- The reservoir characterisation examples in this paper example can be found at:
  https://github.com/NVIDIA/modulus-sym/tree/4c5e6bb9e651bcabb2fee1b2bca75be5bdace056/examples/reservoir_simulation

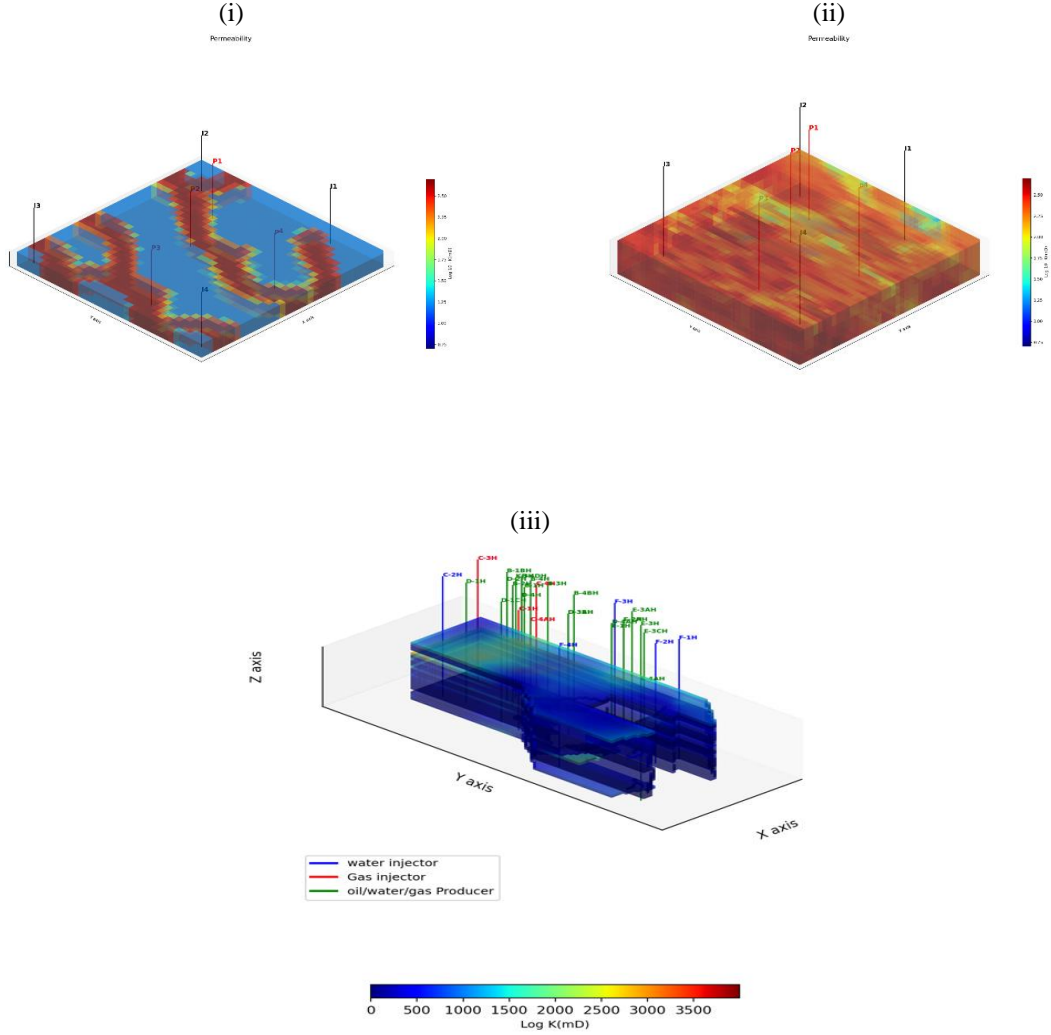

*Figure 3: Cartoon showing the 2 synthetic models and 1 modified real model for the numerical experiments. (1) 2D Channelised bimodal reservoir (ii) 3D Gaussian sandstone reservoir and (iii) Norne Field*

### 3.1 Numerical experiment 1 - 2D

The result for the *PINO* surrogate is shown in Fig. (5), 10,000 training samples (9800 samples for the physics loss without any labelling and 200 samples for the data loss requiring labelling and running our in-house GPU accelerated CFD simulator) were used. The water flows from the injectors (downwards-facing arrows) towards the producers (upwards-facing arrows). The size of the reservoir computational voxel is $nx, ny, nz$ = 33,33,1. Two phases are considered (oil and water) and the wells (4 injectors and 4 producers) are arranged in an "analogous 5- spot pattern" as shown in Fig. 5. The 4 producers well have measurable quantities of oil rate, water rate, and water-cut and are controlled by bottom-hole-pressure. The 4 water injector wells have a measurable quantity of bottom hole pressure (BHP), controlled by injection rates. The reservoir is a sandstone channelised reservoir consisting of 2 lithofacies (sand and shale as shown in Fig. 8). 2,340 days of simulation are simulated. The left column of all panels in Fig. 5 is the responses from the *PINO* surrogate, the middle column is the responses from the in-house finite volume solver with AMG and the right column is the difference between each response. For all panels in Fig. 5, the first row is for the pressure, the second row is for the water saturation and the third row is for the gas saturation.

The root-mean-square-error (RMSE) function for each ensemble member ($i$) is defined as

$$\text{RMSE}(i) = \left( \frac{1}{N} \sum_{k=1}^{N} \sum_{j=1}^{N_{data}^k} \left( \frac{D_{obs}^j(k) - D_{sim,i}^j(k)}{\sigma_{n,j}} \right)^2 \right)^{\frac{1}{2}}$$

Eqn. (29)

$N$: Number of data assimilation time steps where measurements are assimilated (measurement times)
$N_{data}^k$: Number of data collected at each time step $k$
$i$: Ensemble member index
$k$: Time index
$j$: Metric or response (history matched metric or response)
$D_{obs}^j(k)$: Observed data metric for metrics j (Data equivalent in state space ensemble) at time step k.
$D_{sim}^j(k)$: Simulated data from simulator for metrics j (Data equivalent in state space ensemble) at time step k.
$\sigma_{n,j}$: Observed data standard deviation for metrics j (Data equivalent in state space ensemble).

An open-source algorithm SSIM is used to compare the permeability field of the history matched reconstructed permeability realization. SSIM is an image metric quantifier that analyses the visual impact of three identities of an image: structure, contrast, and luminance ($s, c,$ and $l$). For further description of the algorithm, the reader may consult. Explaining SSIM further, a value of 1 means complete similarity while the value of $-1$ indicates complete dissimilarity.
SSIM is defined as:

$$SSIM(x,y) = [l(x,y)]^\alpha \times [c(x,y)]^\beta \times [s(x,y)]^\gamma$$

Eqn. (30a)

where

$$l(x,y) = \frac{2\mu_x\mu_y + C_1}{\mu_x^2 + \mu_y^2 + C_1},$$

Eqn. (30b)

$$c(x,y) = \frac{2\sigma_x\sigma_y + C_2}{\sigma_x^2 + \sigma_y^2 + C_2},$$

Eqn. (30c)

$$s(x,y) = \frac{\sigma_{xy} + C_3}{\sigma_x\sigma_y + C_3}$$

Eqn. (30d)

$\mu_x, \mu_y, \sigma_x\sigma_y$, and $\sigma_{xy}$ are the local means, standard deviations, and cross-covariance for images $x$ (reference image or true model maps), and $y$ (either of the history matched realisations). If $\alpha = \beta = \gamma = 1$, and $C_3 = C_2/2$, $C_1 = (B_1 L)^2, C_2 = (B_2 L)^2$ where $C_1$ and $C_2$ are two variables that stabilises the division with a weak denominator, $L$ is the dynamic range of the value of the pixels, $B_1 = 0.01$ and $B_2 = 0.03$, the index then becomes:

$$SSIM(x,y) = \frac{(2\mu_x\mu_y + C_1)(2\sigma_{xy} + C_2)}{(\mu_x^2 + \mu_y^2 + C_1)(\sigma_x^2 + \sigma_y^2 + C_2)}$$

Eqn. (30e)

$$\phi(SSIM) = abs(1 - SSIM)$$

Eqn. (30f)

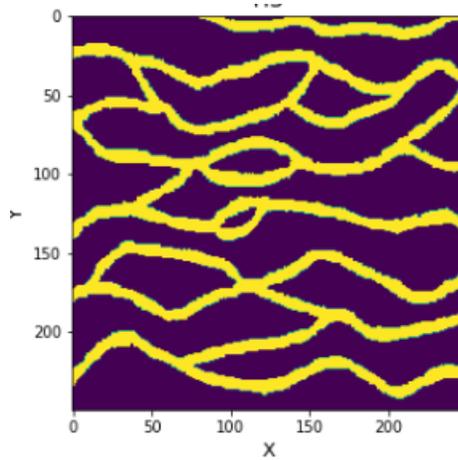

***Figure 4:***  *Training image used to get the prior pdf for the permeability field for numerical experiment 1.*

### 3.1.1 Forward Modelling

*Table 1: Reference Reservoir synthetic model properties for Numerical experiment 1*

| Property | Value |
|---|---|
| Grid Configuration | $33 \times 33 \times 1$ |
| Grid size | $50 \times 50 \times 20 \ ft$ |
| Well configuration | 8-spot |
| # of producers | 4 |
| # of injectors | 4 |
| Simulation period | 3000 days |
| Integration step length | 100 days |
| # of integration steps | 30 |
| Reservoir depth | 4,000ft |
| Initial Reservoir pressure | 1,000psia |
| Injector's constraint | 500 STB/day |
| Producer's constraint | 100 psia |
| Residual oil saturation | 0.2 |
| Connate water saturation | 0.2 |
| Petro-physical property distribution | Bi-modal permeability/porosity |
| Uncertain parameters to estimate | Distributed permeability, porosity |
| Historical data | $Q_{oil}, Q_{water}, BHP, water-cut$ |

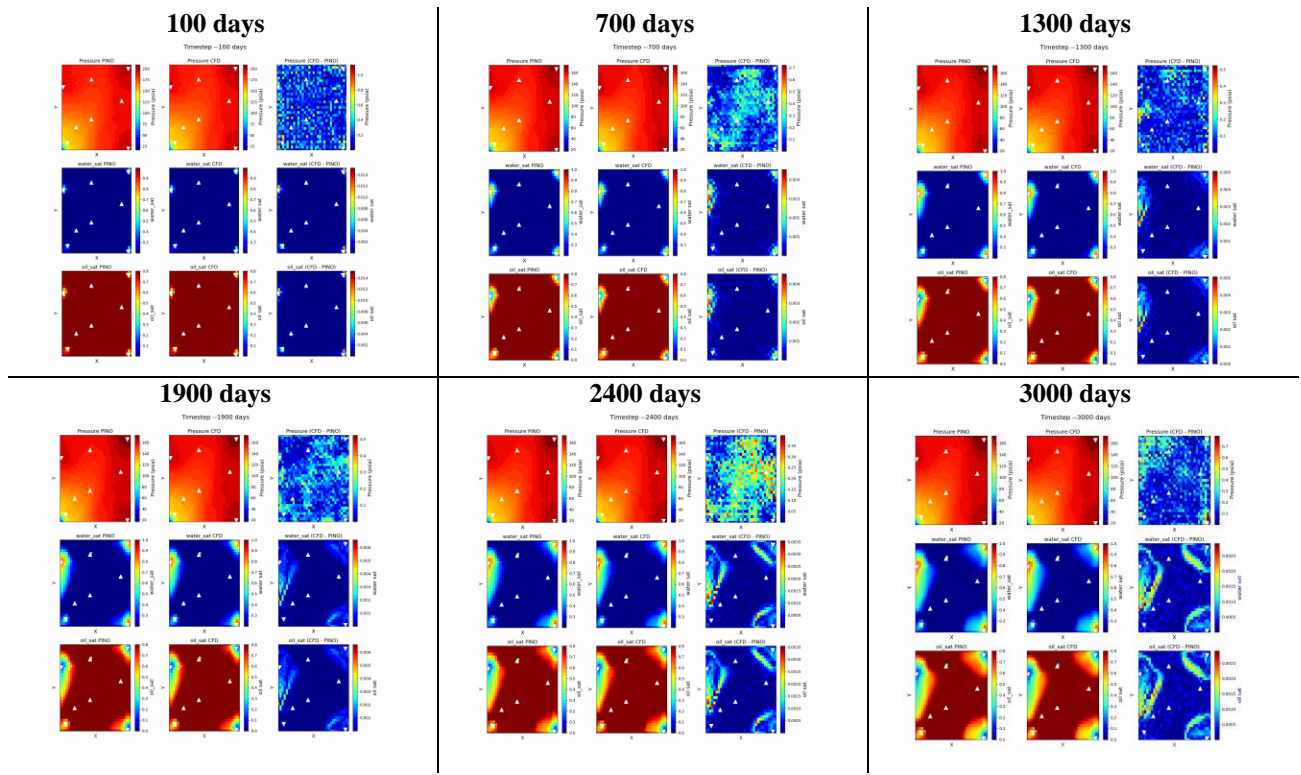

*Figure 5:* *Numerical implementation of Reservoir forward simulation evolution. PINO-based reservoir forwarding (left column), Finite volume method (FVM) based reservoir forwarding (first principle), (middle column) and the difference in magnitudes from both approaches (last column)*

*Table 2: FVM-GPU simulator specifics (single-node + single GPU)*

| Property | Value |
| --- | --- |
| Discretisation | Finite-Volume. (Central in space and forward in time) |
| Solver scheme | Adaptive implicit (Based on CFL number) |
| Pressure solver | V-cycle n stage AMG (PMIS coarsening + Level scheduling + graph matching + colouring) |
| Saturation solver | Left- preconditioned GMRES with ILU (0) |
| Jacobian formation | CSR |
| Boundary conditions | Dirichlet, Neumann |

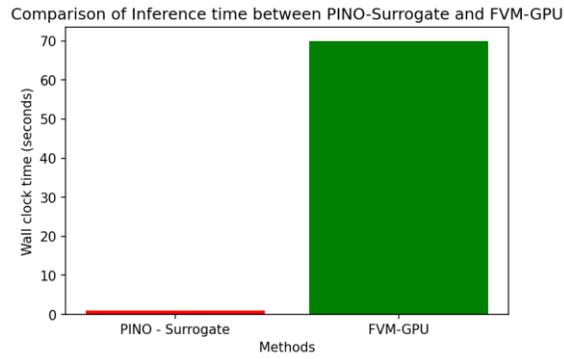

*Figure 6*: *Inference time comparing the FVM-GPU simulator(green) and PINO surrogate (red) for experiment 1*

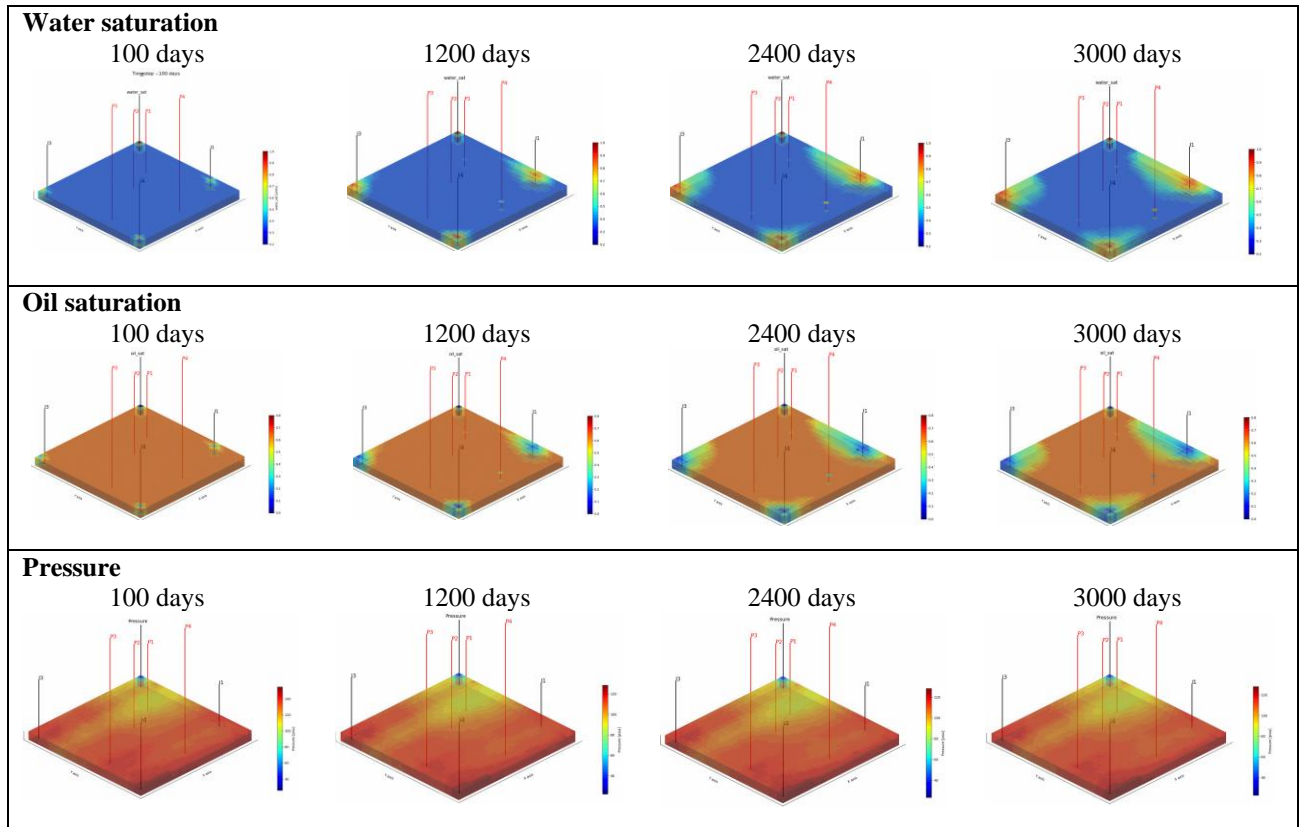

*Figure 7*: *Numerical implementation of Reservoir forward simulation. PINO-based reservoir forwarding showing the water saturation, oil saturation and pressure.*

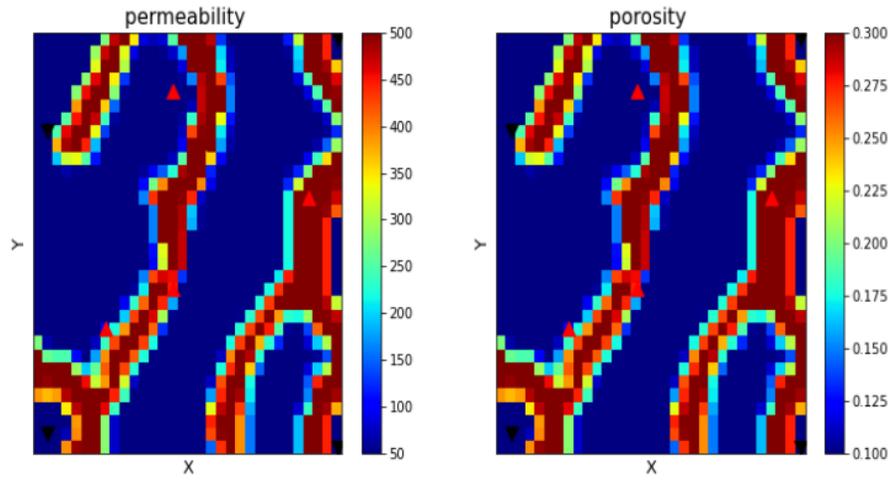

*Figure 8: Synthetic reservoir model used for the experiment 1*

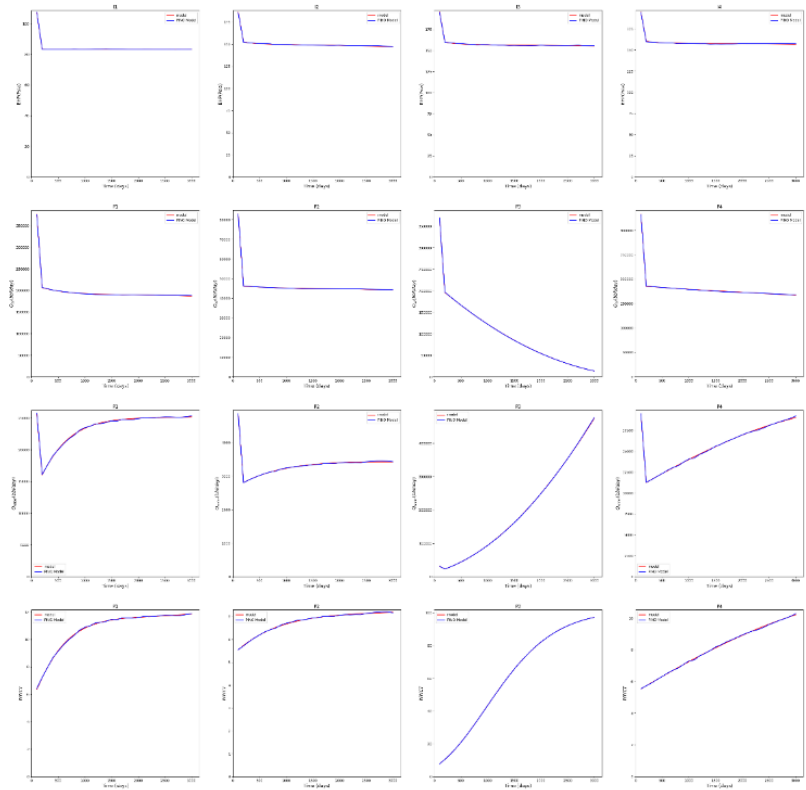

*Figure 9: Production profile comparison. (red) True model, (blue) PINO model. The first row is for the bottom-hole-pressure of well injectors (I1-I4), the second row is for the oil rate production for the well producers (P1-P4), the third row is for the water rate production for the well producers (P1-P4) and the last row is for the water cut ratio of the 4 well producers (P1-P4)*

### 3.1.2 Inverse modelling

The parameter to recover here is,

$$u = \sum_{j=1}^{J} \begin{bmatrix} K \\ \varphi \end{bmatrix}^{j}$$

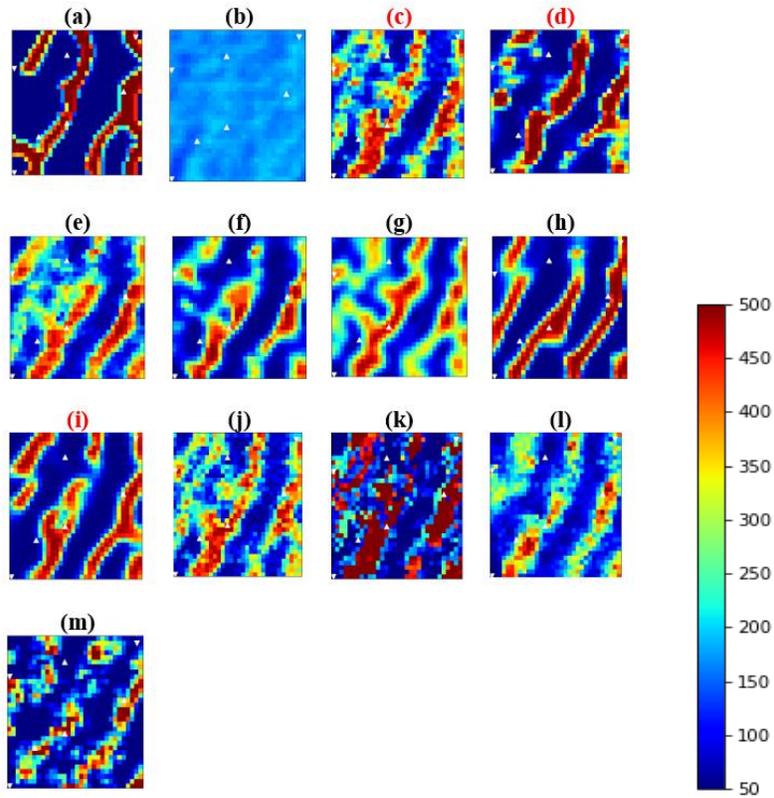

*Figure 10:* *Comparison of permeability field reconstruction (a) True Model (b) Prior, (c) Posterior – aREKI + KSVD + OMP, (d) Posterior – aREKI + CCR, (e) Posterior – aREKI + KMEANS, (f) Posterior – aREKI + Convolution autoencoder, (g) Posterior – aREKI + DCT, (h) Posterior – aREKI + Denoising Convolution autoencoder, (i) Posterior – aREKI + Variational Convolution Auto Encoder, (j) Posterior – Vanilla aREKI, (k) Posterior – aREKI + PCA, (l) Posterior aREKI + Level set (m) Posterior – aREKI SVM*

In Fig. 10, we compare the developed *PINO*-aREKI with various parameterisation methods with the 3 selected methods. Which is characterisation with KSVD + OMP [12], characterisation with CCR [13 & 53] and characterisation with VCAE. We notice the improved geological realism from these 3 methods when compared to other methods.

*Table 3: RMS & SSIM cost function of MAP models from different methods.*

| Model | $\phi(SSIM)$ | RMS function | Overall cost function [$\phi(SSIM)$ + RMS function] |
|---|---|---|---|
| aREKI + KSVD + OMP | 0.129 | 0.044 | 0.173 |
| aREKI + CCR | 0.131 | 0.046 | 0.177 |
| aREKI + KMEANS | 0.175 | 0.081 | 0.256 |
| aREKI + Convolution autoencoder | 0.134 | 0.053 | 0.187 |
| aREKI + DCT | 0.142 | 0.058 | 0.2 |
| aREKI + Denoising Convolution autoencoder, | 0.135 | 0.051 | 0.186 |
| aREKI + Variational Convolution Auto Encoder | 0.124 | 0.044 | 0.168 |
| Vanilla aREKI | 0.234 | 0.081 | 0.315 |
| aREKI + PCA | 0.215 | 0.077 | 0.292 |
| aREKI + Level set | 0.146 | 0.069 | 0.215 |

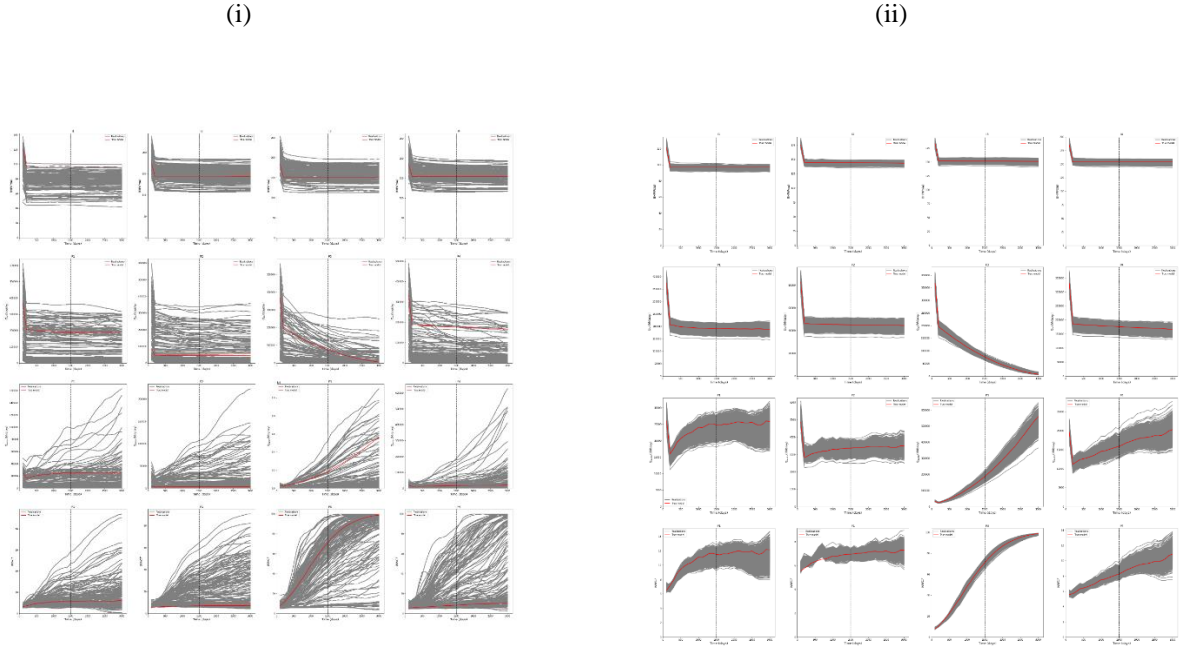

(i)      (ii)

*Figure 11: Production profile comparison for the prior ensemble (i) and Production profile comparison for the posterior ensemble(ii). (red) True model, (grey) Ensemble. The first row is for the bottom-hole-pressure of well injectors (I1-I4), the second row is for the oil rate production for the well producers (P1-P4), the third row is for the water rate production for the well producers (P1-P4) and the last row is for the water cut ratio of the 4 well producers (P1-P4). The left of the dash vertical line is used for assimilation while the right of this line is used for*

*prediction.*

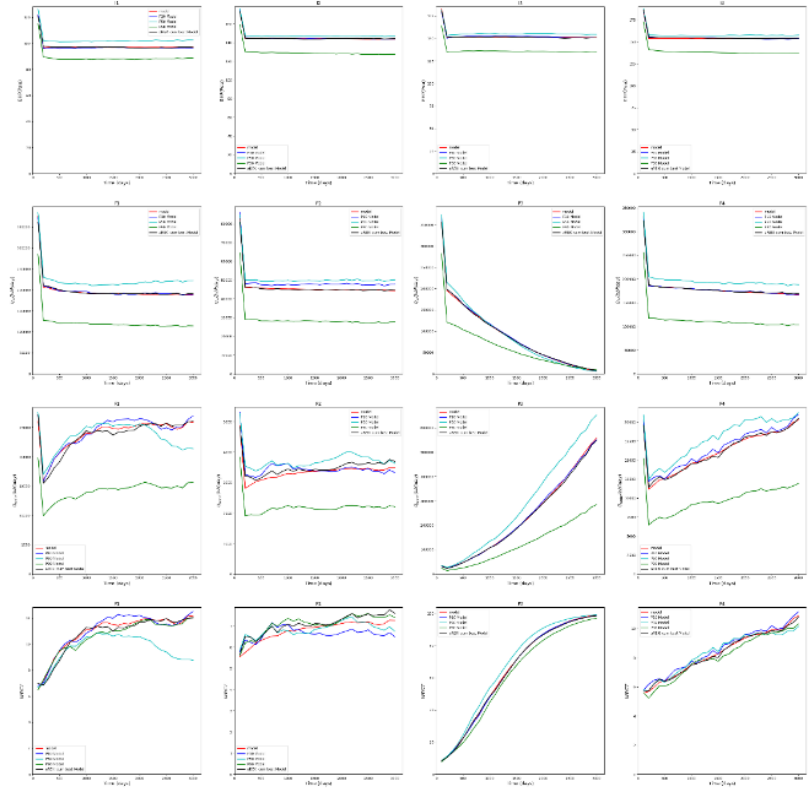

*Figure 12: P10-P50-P90 Production profile comparison for the posterior ensemble. (red) True model, (Blue) P10 model, (cyan) P50 model, (green) P90 model. The first row is for the bottom-hole-pressure of well injectors (I1-I4), the second row is for the oil rate production for the well producers (P1-P4), third row is for the water rate production for the well producers (P1-P4) and the last row is for the water cut ratio of the 4 well producers (P1-P4). The left of the dash vertical line is used for assimilation while the right of this line is used for prediction.*

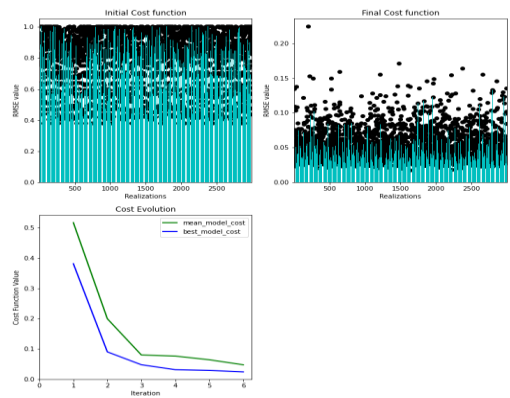

*Figure 13: Cost function evolution. (Top-left) the initial cost function for the initial ensemble, (top-right) Final cost function for the posterior ensemble. (Bottom-left) Cost function evolution for both the MAP model and the best*

*model. Notice the number of iterations (less than 10) necessary for convergence to the posterior density.*

Fig. 13 shows the elegance of the developed *a*REKI, we can sample the posterior density with the fewest iteration possible.

***Table 4:*** *RMS cost function evolution of Best 7 Realisations with the proposed method.*

| Realisation | RMS value of prior model | RMS value of posterior model |
|---|---|---|
| 6 | 0.81 | 0.071 |
| 17 | 0.77 | 0.062 |
| 30 | 0.83 | 0.042 |
| 33 | 0.93 | 0.06 |
| 93 | 0.88 | 0.051 |
| 54 | 0.89 | 0.053 |
| 41 | 0.93 | 0.044 |

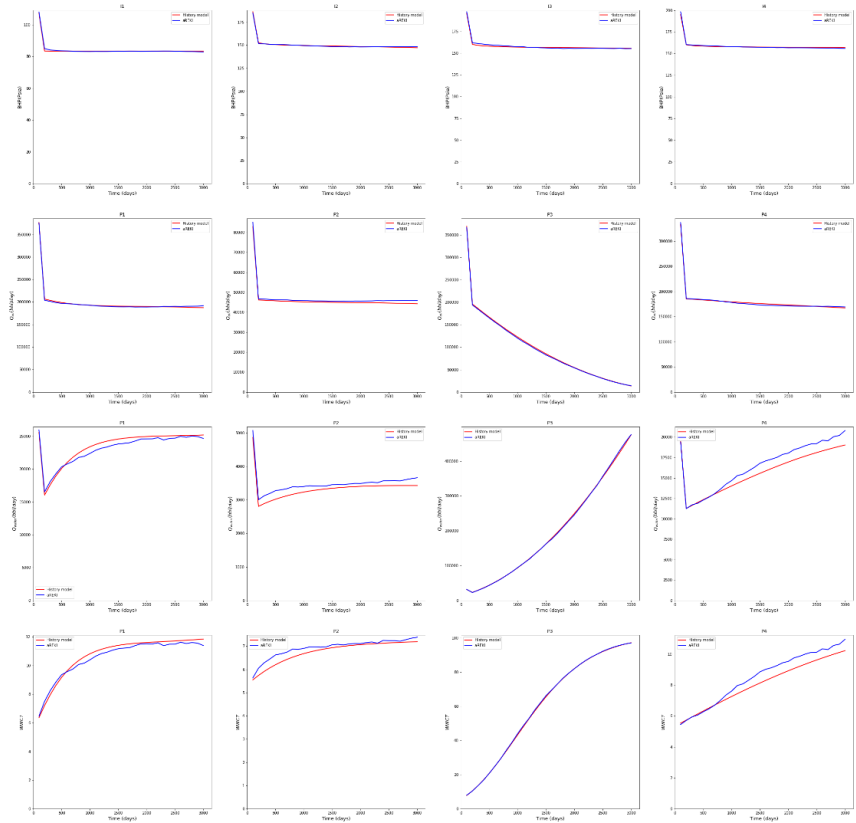

***Figure 14****: Production profile comparison for MAP model. (red) True model, (Blue) MAP model after aREKI + VCAE+ CCR. The first row is for the bottom-hole-pressure of well injectors (I1-I4), the second row is for the oil rate production for the well producers (P1-P4), the third row is for the water rate production for the well producers (P1-P4) and the last row is for the water cut ratio of the 4 well producers (P1-P4)*

## 3.2 Numerical experiment 2 - 3D

The result for the *PINO* surrogate for experiment 2 is shown in Fig. (15-17), 10,000 training samples (9800 samples for the physics loss without any labelling and 200 samples for the data loss requiring labelling and running our in-house GPU accelerated CFD simulator) were used. The water flows from the injectors (downwards-facing arrows) towards the producers (upwards-facing arrows). The size of the reservoir computational voxel is $nx, ny, nz = 40,40,3$. Two phases are considered (oil and water) and the wells (4 injectors and 4 producers) are arranged in an "analogous 5- spot pattern". The 4 producers well have measurable quantities of oil rate, water rate, and water-cut and are controlled by bottom-hole-pressure. The 4 water injector wells have a measurable quantity of bottom hole pressure (BHP), controlled by injection rates. The reservoir is a sandstone Gaussian reservoir. 2,340 days of simulation are simulated. The left column of Fig. 15-16 is the responses from the *PINO* surrogate, the middle column is the responses from the finite volume solver with AMG and the right column is the difference between each response.

*Table 5: Reference Reservoir synthetic model properties for Numerical experiment 2*

| Property | Value |
|---|---|
| Grid Configuration | $40 \times 40 \times 3$ |
| Grid size | $50 \times 50 \times 20\ ft$ |
| Well configuration | 8-spot |
| # of producers | 4 |
| # of injectors | 4 |
| Simulation period | 3000 days |
| Integration step length | 100 days |
| # of integration steps | 30 |
| Reservoir depth | 4,000ft |
| Initial Reservoir pressure | 1,000psia |
| Injector's constraint | 500 STB/day |
| Producer's constraint | 100 psia |
| Residual oil saturation | 0.2 |
| Connate water saturation | 0.2 |
| Petro-physical property distribution | Gaussian permeability/porosity |
| Uncertain parameters to estimate | Distributed permeability, porosity |

### 3.2.1 Forward modelling

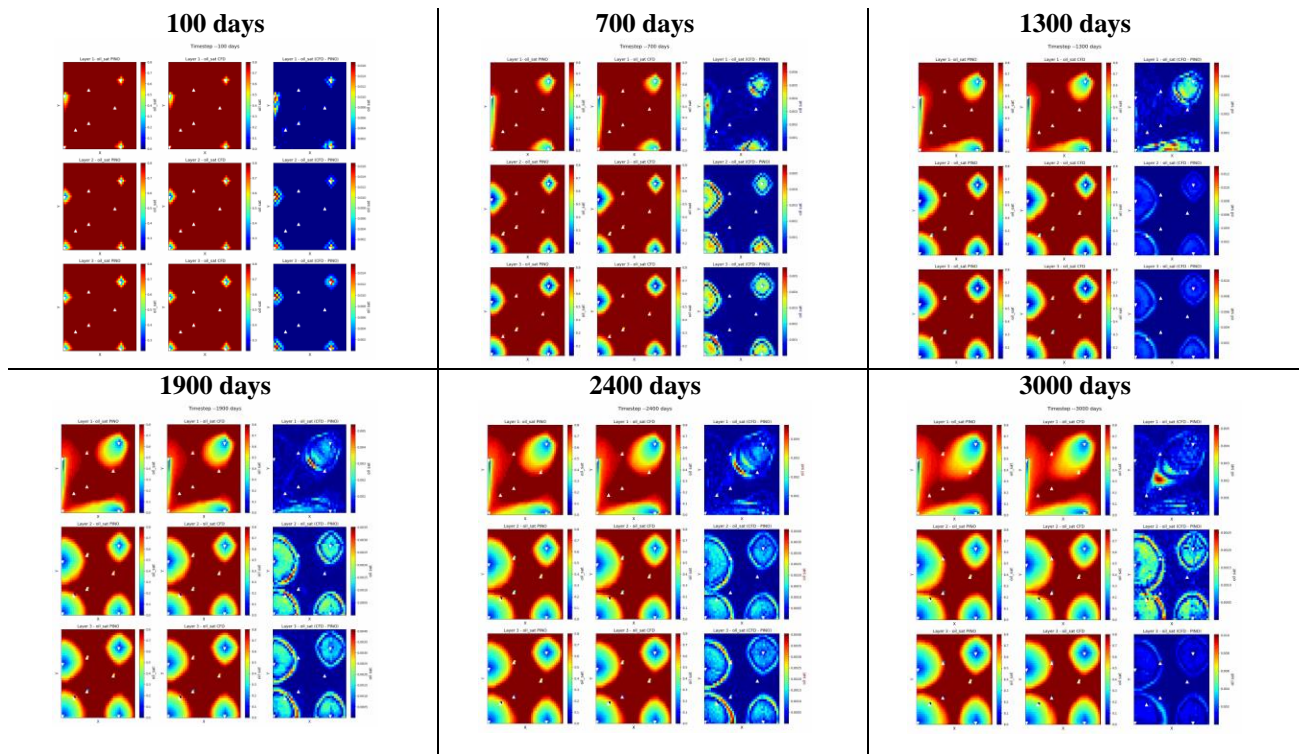

*Figure 15: Oil saturation evolution Numerical implementation of Reservoir forward simulation. PINO-based reservoir forwarding (left column), Finite volume method (FVM) based reservoir forwarding (first principle), (middle column) and the difference in magnitudes from both approaches (last column)*

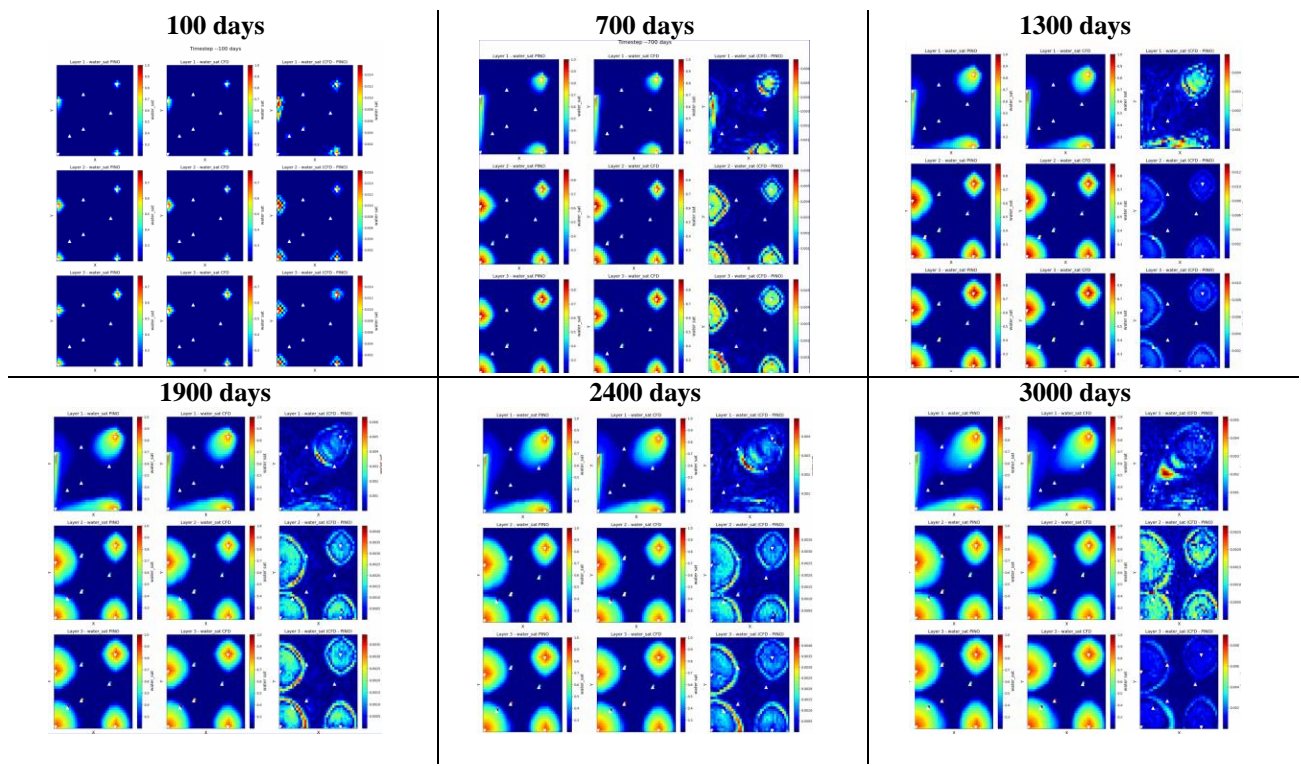

*Figure 16: water saturation evolution Numerical implementation of Reservoir forward simulation. PINO-based reservoir forwarding (left column), Finite volume method (FVM) based reservoir forwarding (first principle),*

*(middle column) and the difference in magnitudes from both approaches (last column)*

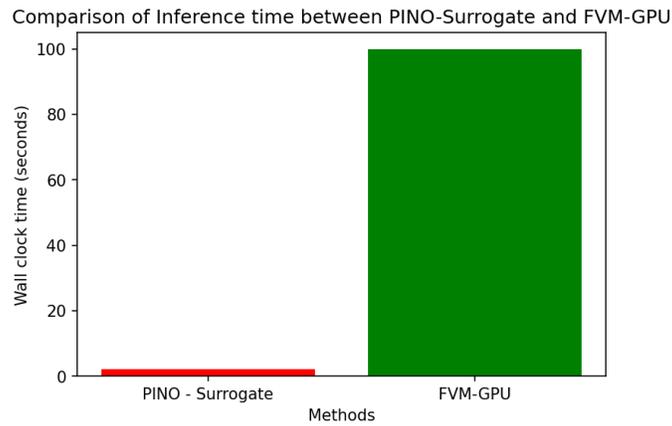

*Figure 17: Inference time comparing the FVM-GPU simulator(green) and PINO surrogate (red) for experiment.*

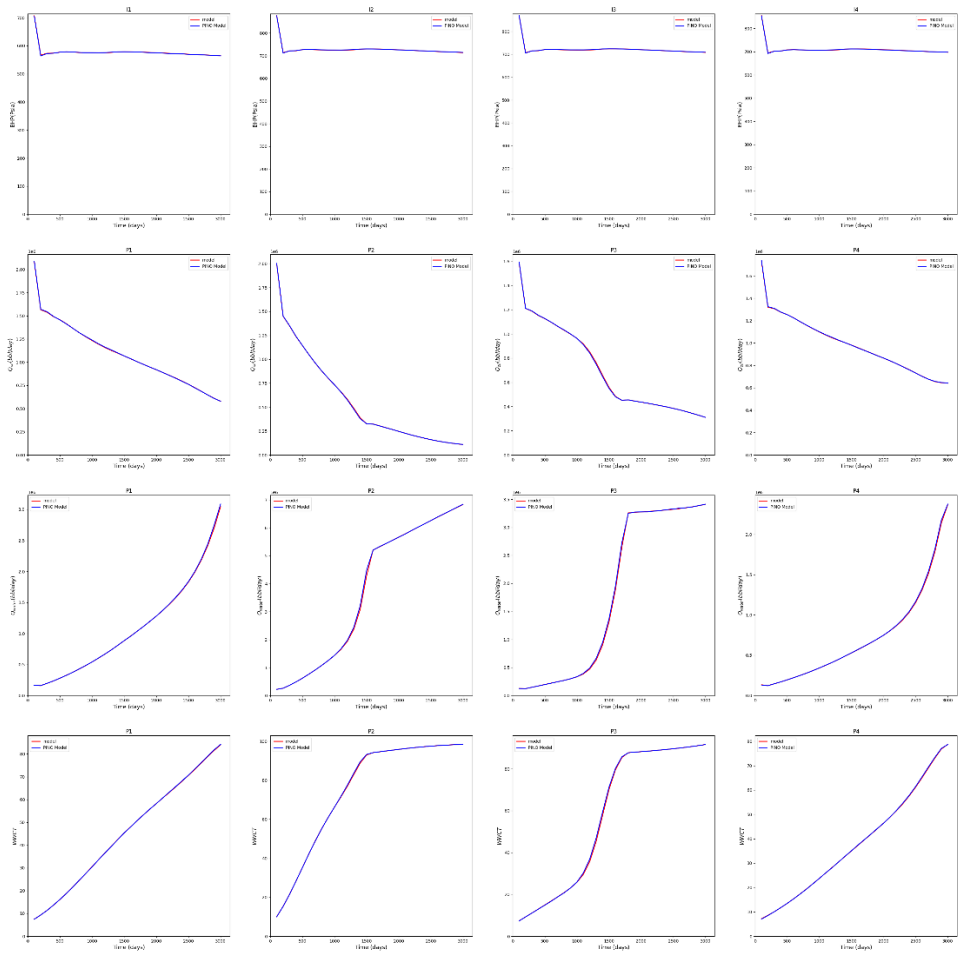

*Figure 18: Production profile comparison. (red) True model, (blue) PINO model. First row is for the bottom-hole-pressure of well injectors (I1-I4), second row is for the oil rate production for the well producers (P1-P4), third row is for the water rate production for the well producers (P1-P4) and the last row is for the water cut ratio of the 4 well producers (P1-P4)*

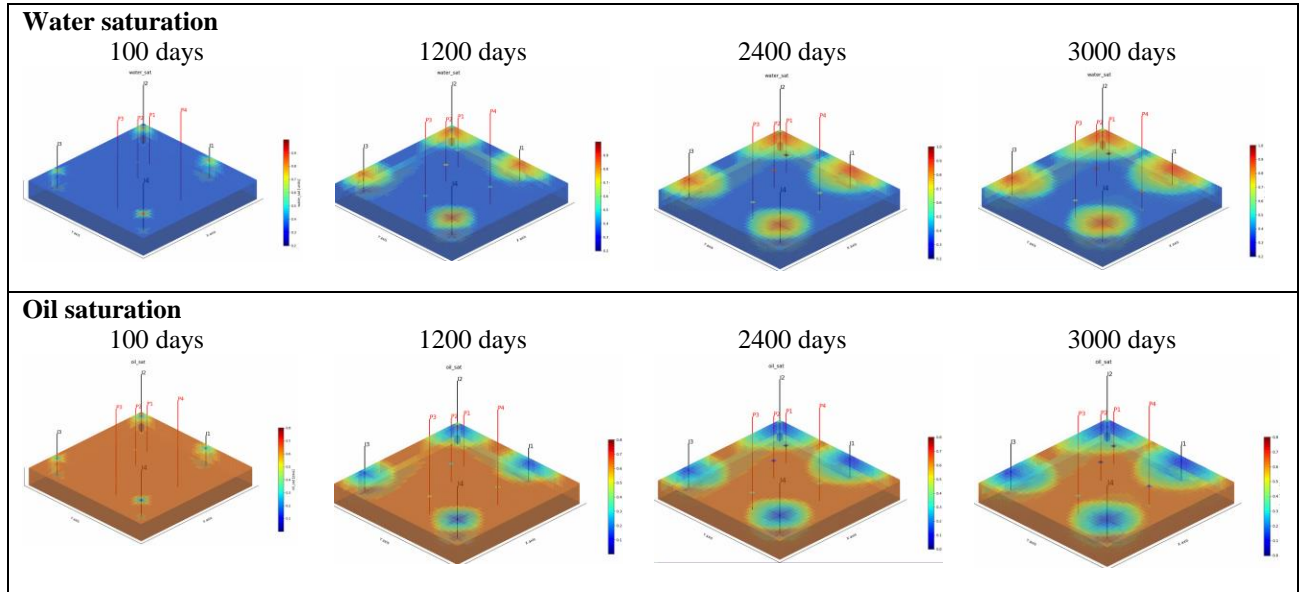

**Figure 19:** Numerical implementation of Reservoir forward simulation. *PINO*-based reservoir forwarding. Showing the water saturation and oil saturation.

### 3.2.2 Inverse modelling

The parameter to recover here is,

$$u = \sum_{j=1}^{J} \begin{bmatrix} K \\ \varphi \end{bmatrix}^{j}$$

(i) (ii)

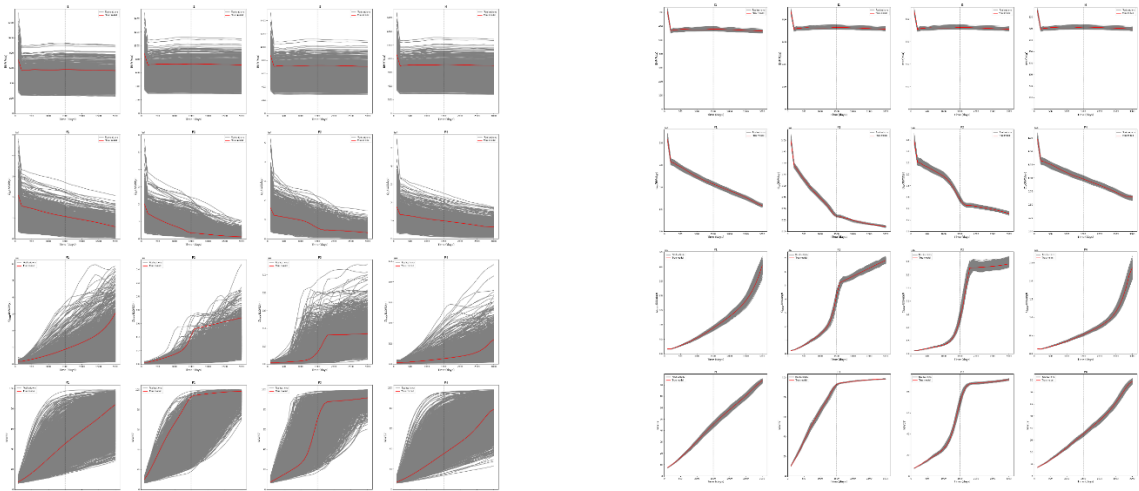

***Figure 20:*** *Production profile comparison for prior ensemble (i) and posterior ensemble (ii). (red) True model, (grey) Ensemble. First row is for the bottom-hole-pressure of well injectors (I1-I4), second row is for the oil rate production*

*for the well producers (P1-P4), third row is for the water rate production for the well producers (P1-P4) and the last row is for the water cut ratio of the 4 well producers (P1-P4). Left of the dash vertical line is used for assimilation while right of this line is used for prediction.*

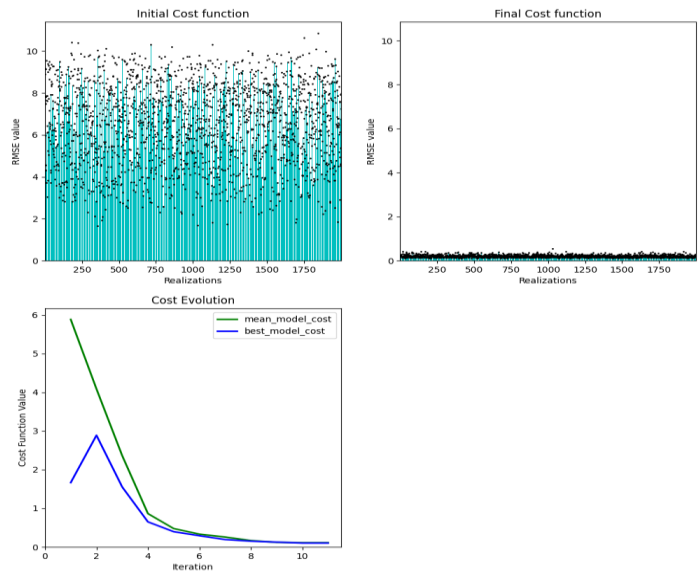

*Figure 21: Cost function evolution. (Top -left) prior ensemble RMSE cost, (Top-right) posterior ensemble RMSE cost, (Bottom-left) RMSE cost evolution between the MAP model (blue) and the MLE model (green)*

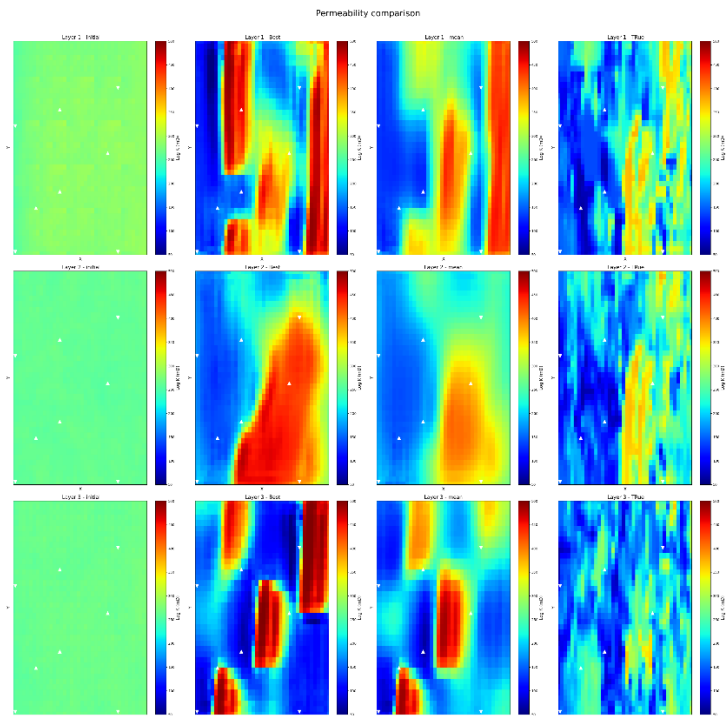

*Figure 22: Comparison of permeability field reconstruction (1st column) prior, (2nd column) MLE estimate (posterior) (third column) MAP estimate (posterior) and (4th-column) True Model. The method used is the – aREKI + Convolution autoencoder.*

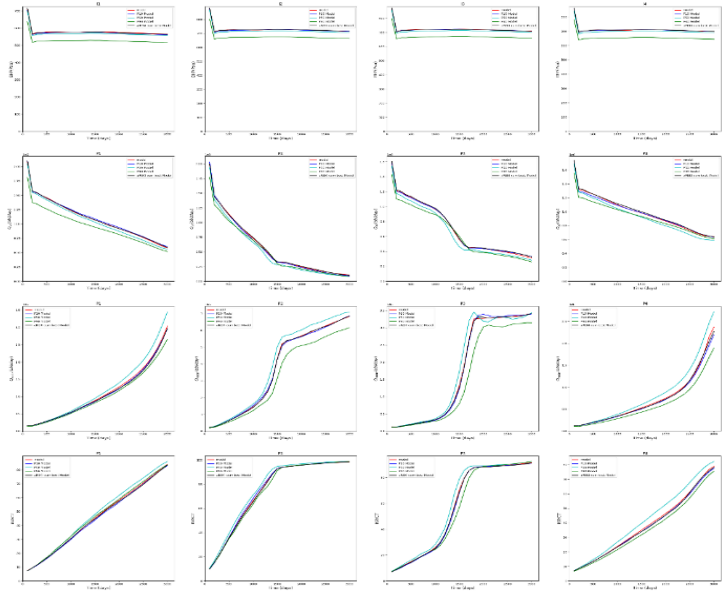

*Figure 23*: *P10-P50-P90 Production profile comparison for the posterior ensemble. (red) True model, (Blue) P10 model, (cyan) P50 model, (green) P90 model. The first row is for the bottom-hole-pressure of well injectors (I1-I4), the second row is for the oil rate production for the well producers (P1-P4), third row is for the water rate production for the well producers (P1-P4) and the last row is for the water cut ratio of the 4 well producers (P1-P4). The left of the dash vertical line is used for assimilation while the right of this line is used for prediction.*

## 3.3 Numerical experiment 3 - *Norne*

The result for the *PINO* surrogate for the *Norne* field is shown in Fig.24 (a-d), The *Norne* reservoir model is based on a real field black-oil model for an oil field in the Norwegian Sea [62,63]. The grid is a faulted corner-point grid, with heterogenous and anisotropic permeability 100 training samples for the data loss requiring labelling and running the OPM-Flow simulator [61] were used. The water flows from the injectors (downwards-facing arrows) towards the producers (upwards-facing arrows). The size of the reservoir computational voxel is $nx, ny, nz = 46,112,22$. Three phases are considered (oil, gas, and water) and the wells (9 water injectors, 4 gas injectors and 22 producers) are configured. The 22 producers well have measurable quantities of oil rate, water rate, and gas rate and are controlled by bottom-hole-pressure. The 4 water injector wells have a measurable quantity of bottom hole pressure (BHP), controlled by injection rates. The reservoir is a sandstone Gaussian reservoir. 3,298 days of simulation are simulated. 100 realizations of the *Norne* model that include variations in porosity, rock permeabilities, and fault transmissibility multipliers are used to generate the training and test data.

The left column of Fig. 24(a-d) is the responses from the *PINO* surrogate, the middle column is the responses from the finite volume solver -Flow and the right column is the difference between each response.

*Table 5: Reference Reservoir synthetic model properties for Norne*

| Property | Value |
| --- | --- |
| Grid Configuration | $46 \times 112 \times 22$ |
| Grid size | $50 \times 50 \times 20\ ft$ |
| # of producers | 22 |
| # of injectors | 13 |
| Simulation period | 3,298 days |
| Integration step length | 100 days |
| # of integration steps | 30 |
| Reservoir depth | 4,000ft |
| Initial Reservoir pressure | 1,000psia |
| Injector's constraint | 500 STB/day |
| Producer's constraint | 100 psia |
| Residual oil saturation | 0.2 |
| Connate water saturation | 0.2 |
| Petro-physical property distribution | Gaussian permeability/porosity |
| Uncertain parameters to estimate | Distributed permeability, porosity, Fault transmissibility multiplier |

### 3.3.1 Forward modelling

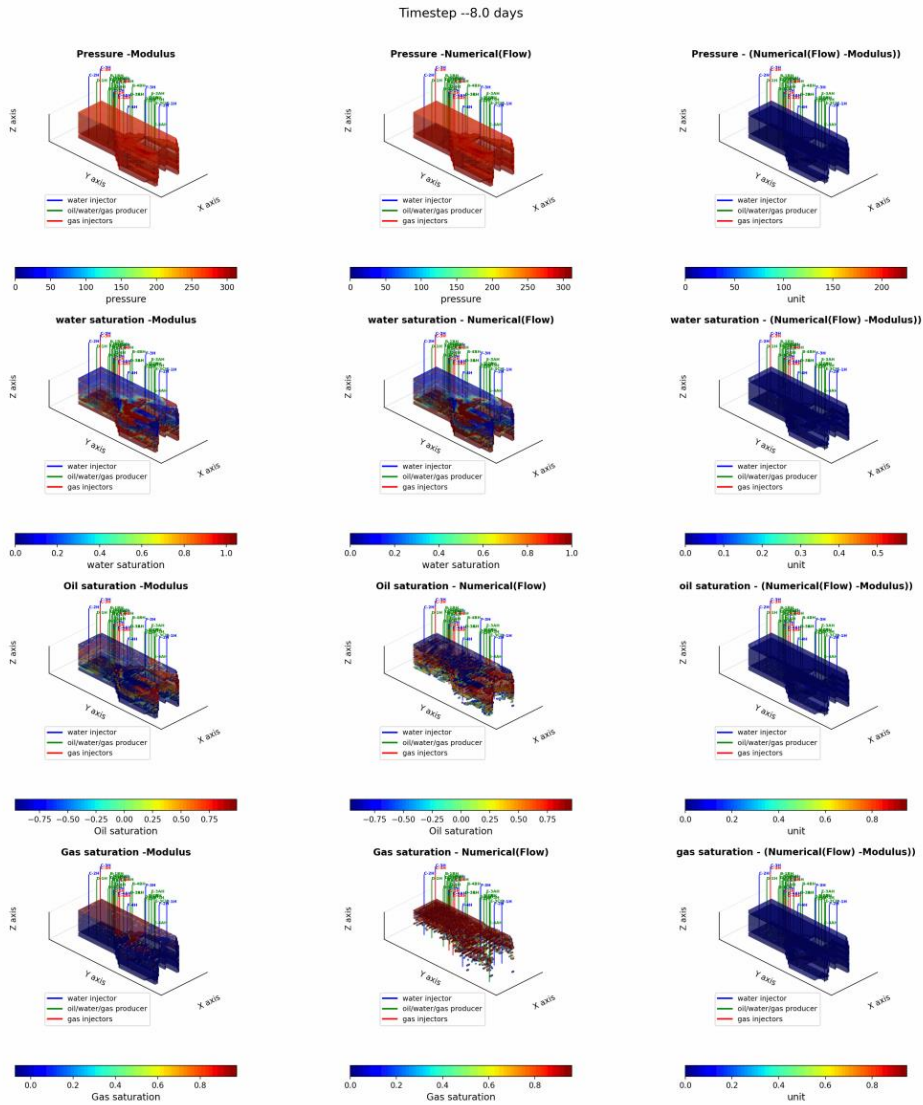

***Figure 24(a):*** *Forwarding of the Norne Field. $N_x = 46$, $N_y = 112$, $N_z = 22$. At Time = 8 days. Dynamic properties comparison between the pressure, water saturation, oil saturation and gas saturation field between NVIDIA Modulus's PINO surrogate (left column), Flow reservoir simulator (middle column) and the difference between both approaches (last column). They are 22 oil/water/gas producers (green), 9 water injectors (blue) and 4 gas injectors) red. We can see good concordance between the surrogate's prediction and the numerical reservoir simulator (Flow)*

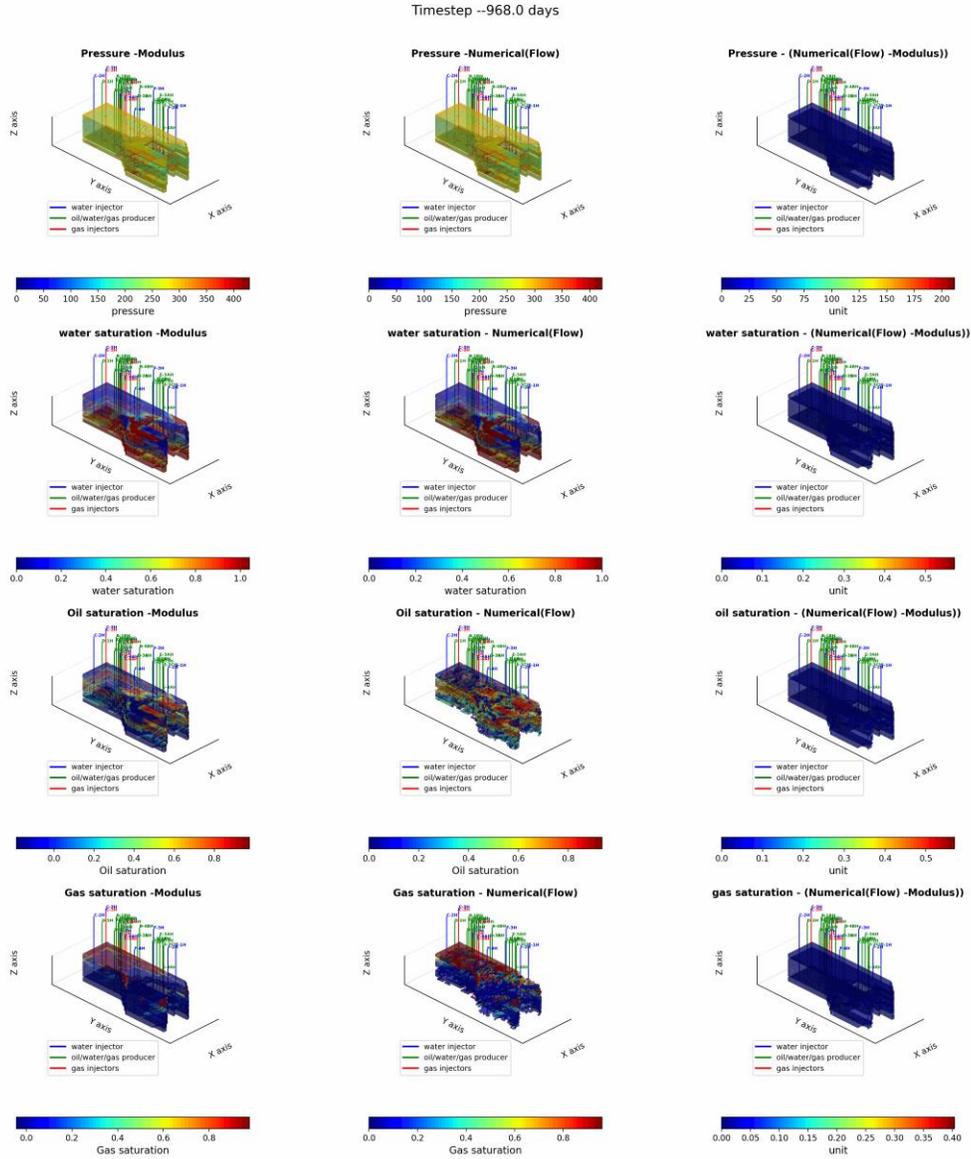

***Figure 24(b):*** *Forwarding of the Norne Field. $N_x = 46$, $N_y = 112$, $N_z = 22$. At Time = 968 days. Dynamic properties comparison between the pressure, water saturation, oil saturation and gas saturation field between NVIDIA Modulus's PINO surrogate (left column), Flow reservoir simulator (middle column) and the difference between both approaches (last column). They are 22 oil/water/gas producers (green), 9 water injectors (blue) and 4 gas injectors) red. We can see good concordance between the surrogate's prediction and the numerical reservoir simulator (Flow)*

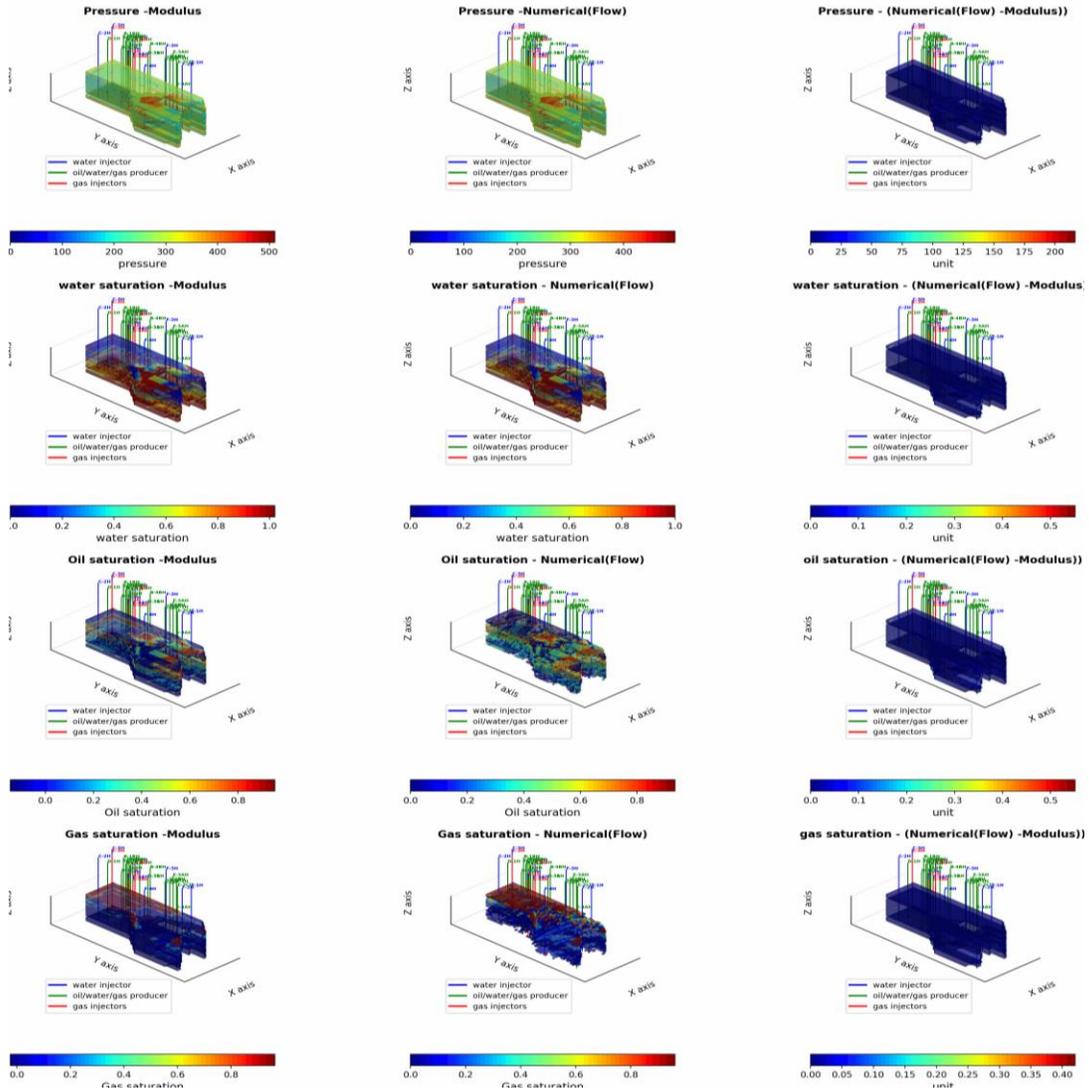

***Figure 24(c):*** *Forwarding of the Norne Field. $N_x = 46$, $N_y = 112$, $N_z = 22$. At Time = 2104 days. Dynamic properties comparison between the pressure, water saturation, oil saturation and gas saturation field between NVIDIA Modulus's PINO surrogate (left column), Flow reservoir simulator (middle column) and the difference between both approaches (last column). They are 22 oil/water/gas producers (green), 9 water injectors (blue) and 4 gas injectors) red. We can see good concordance between the surrogate's prediction and the numerical reservoir simulator (Flow)*

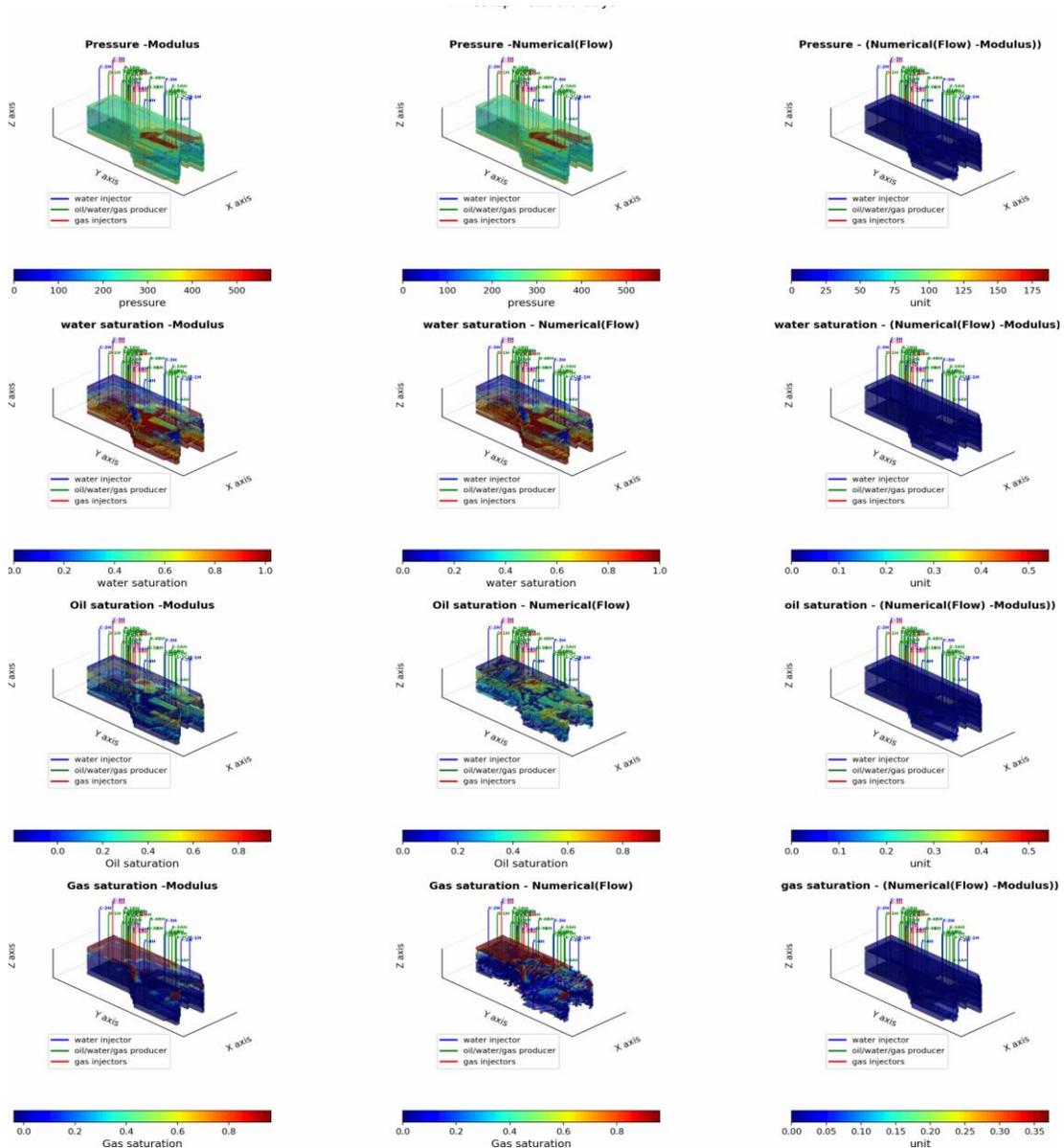

***Figure 24(d):*** *Forwarding of the Norne Field. $N_x = 46$, $N_y = 112$, $N_z = 22$. At Time = 3298 days. Dynamic properties comparison between the pressure, water saturation, oil saturation and gas saturation field between NVIDIA Modulus's PINO surrogate (left column), Flow reservoir simulator (middle column) and the difference between both approaches (last column). They are 22 oil/water/gas producers (green), 9 water injectors (blue) and 4 gas injectors) red. We can see good concordance between the surrogate's prediction and the numerical reservoir simulator (Flow)*

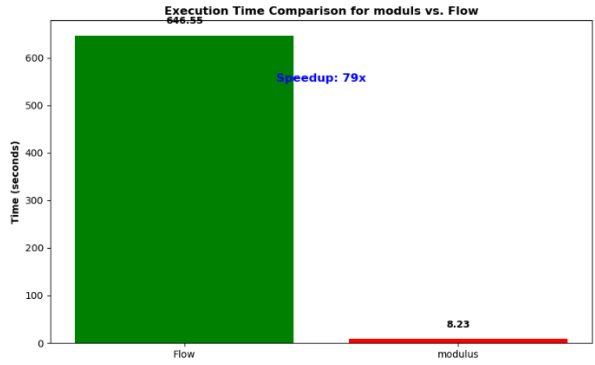

***Figure 25:*** *Forwarding of the Norne Field. $N_x = 46$ , $N_y = 112$ , $N_z = 22$ =46. **(a)** Speed-up using the **PINO-RES-SIM** machine*

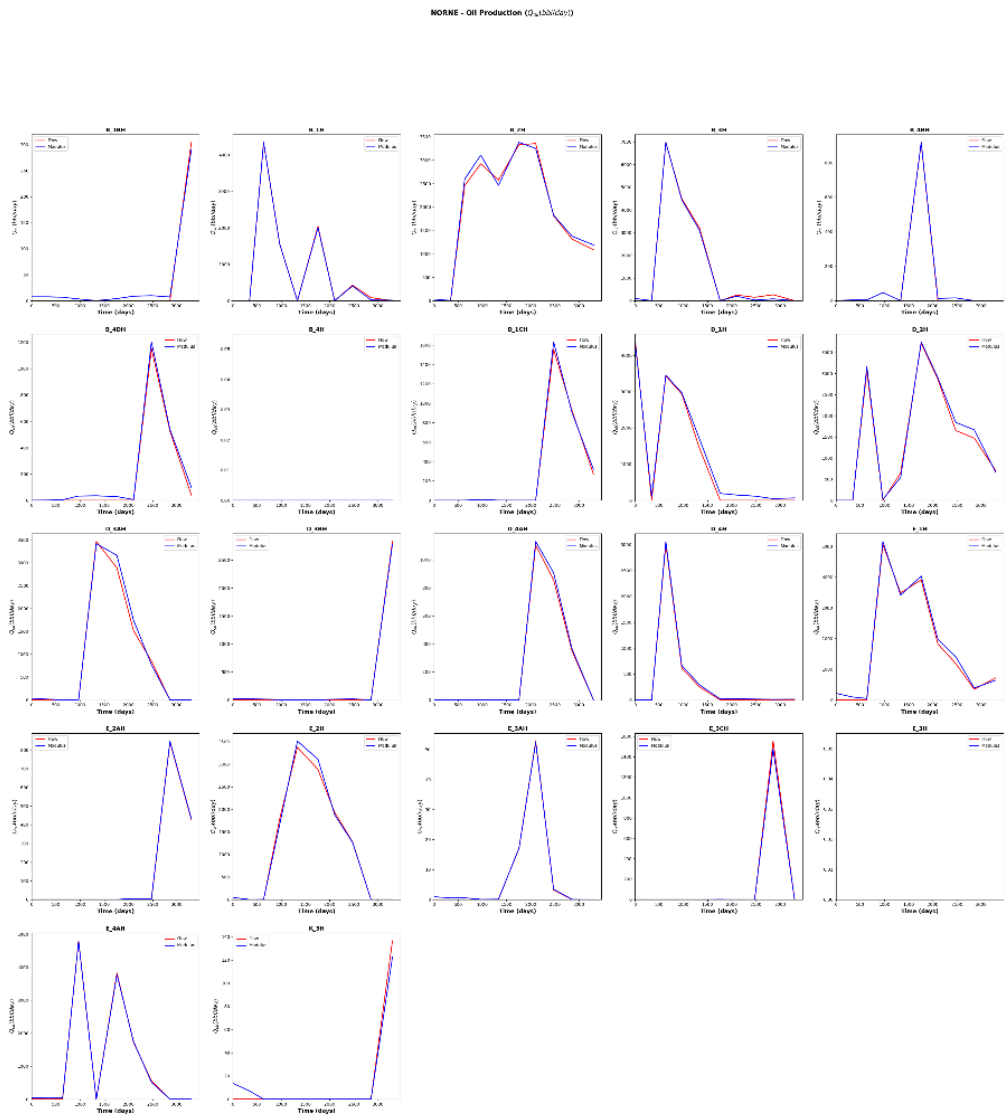

***Figure 26(a):*** *Forwarding of the Norne Field Comparison between using the. (red) True model, (blue) PINO-RES-SIM model. Outputs from peaceman machine for oil production rate,$Q_o \frac{(stb)}{(day)}$ from the 22 producers. They are 22 oil/water/gas producers, 9 water injectors and 4 gas injectors.*

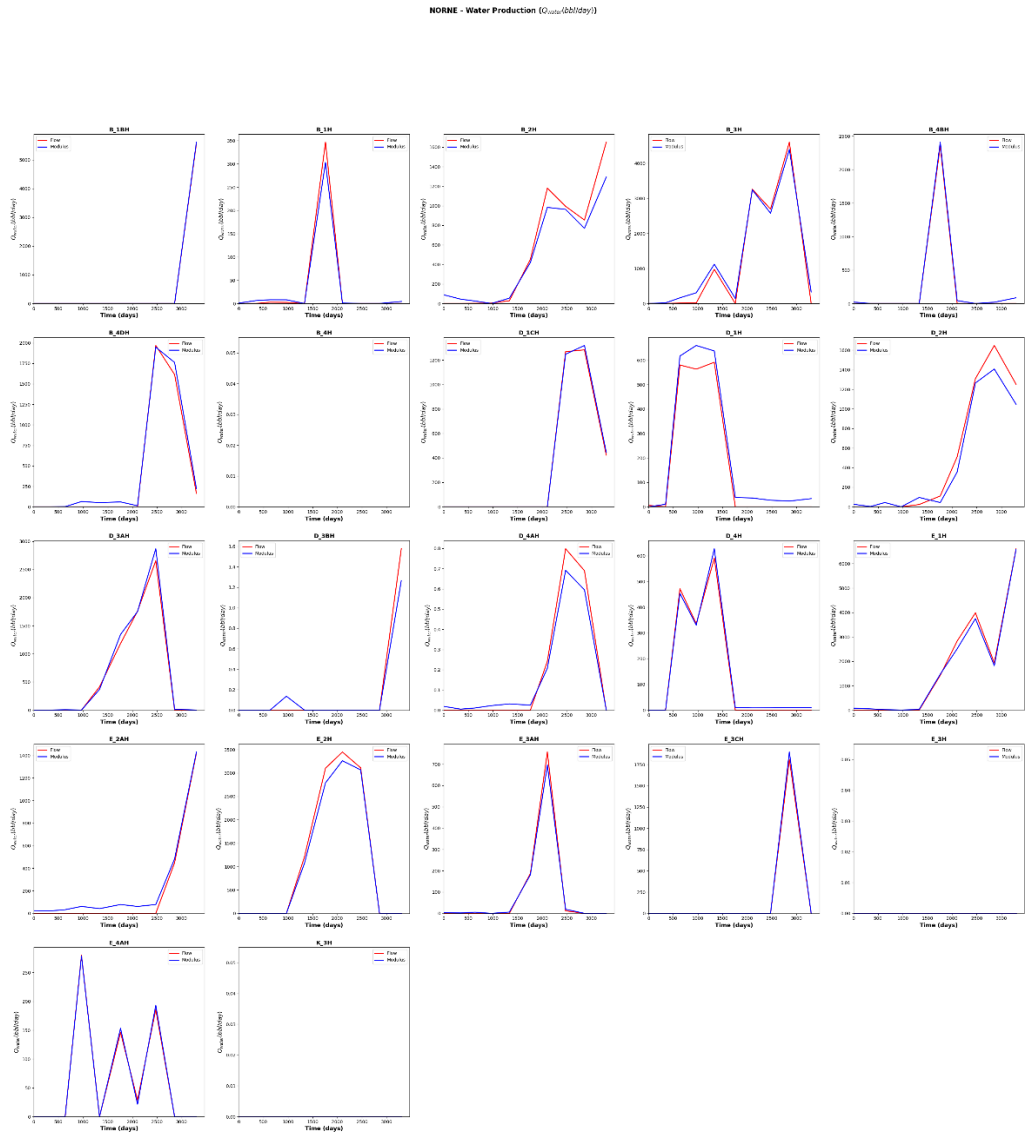

***Figure 26(b):*** *Forwarding of the Norne Field Comparison between using the. (red) True model, (blue) PINO-RES-SIM model. Outputs from peaceman machine for water production rate,$Q_w \frac{(stb)}{(day)}$ from the 22 producers. They are 22 oil/water/gas producers, 9 water injectors and 4 gas injectors.*

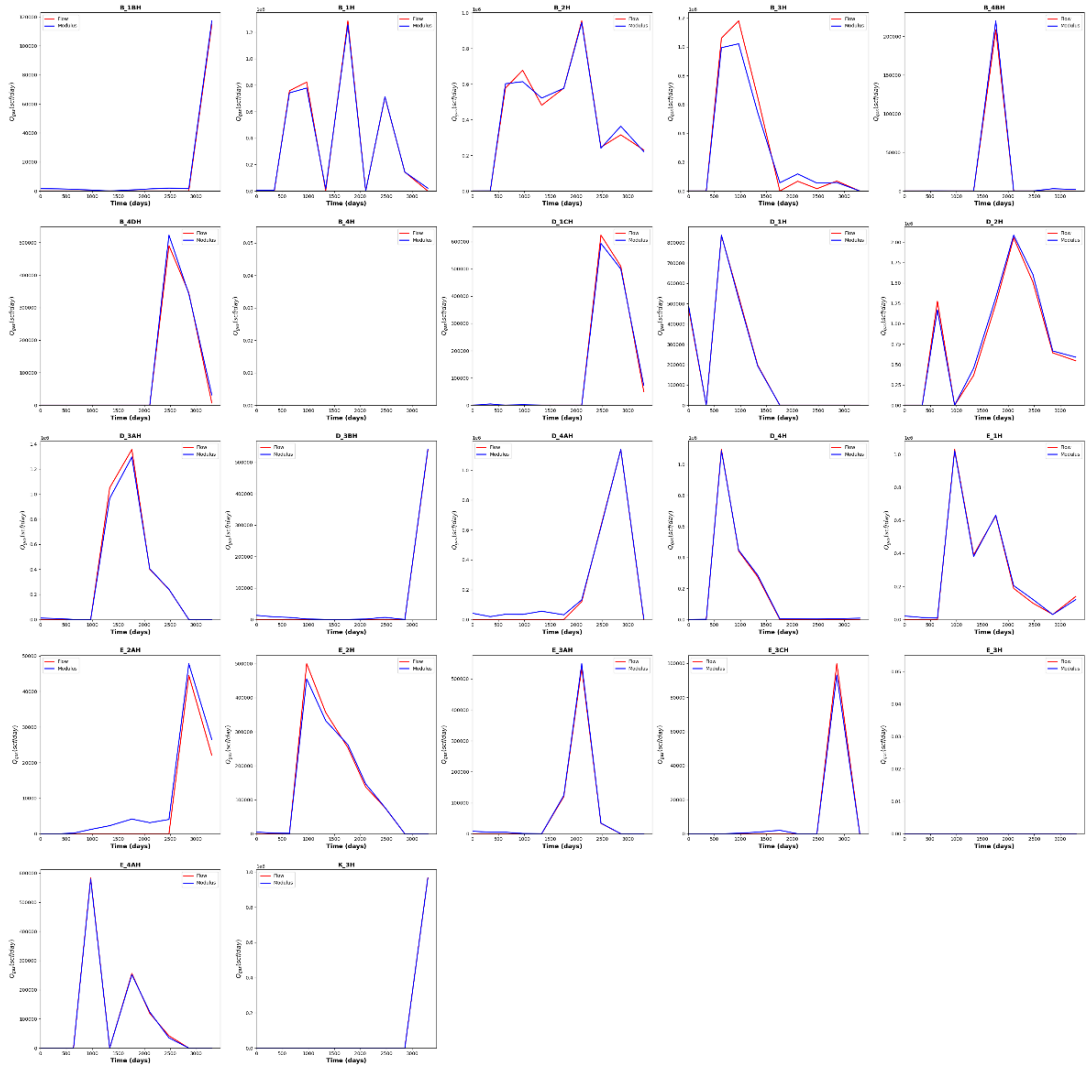

*Figure 26(c): Forwarding of the Norne Field Comparison between using the. (red) True model, (blue) PINO-RES-SIM model. Outputs from peaceman machine for oil production rate, $Q_g \frac{(scf)}{(day)}$ from the 22 producers. They are 22 oil/water/gas producers, 9 water injectors and 4 gas injectors.*

### 3.3.2 Inverse modelling

The parameter to recover here is represented by,

$$u = \sum_{j=1}^{J} \begin{bmatrix} K \\ \varphi \\ FTM \end{bmatrix}^{j}$$

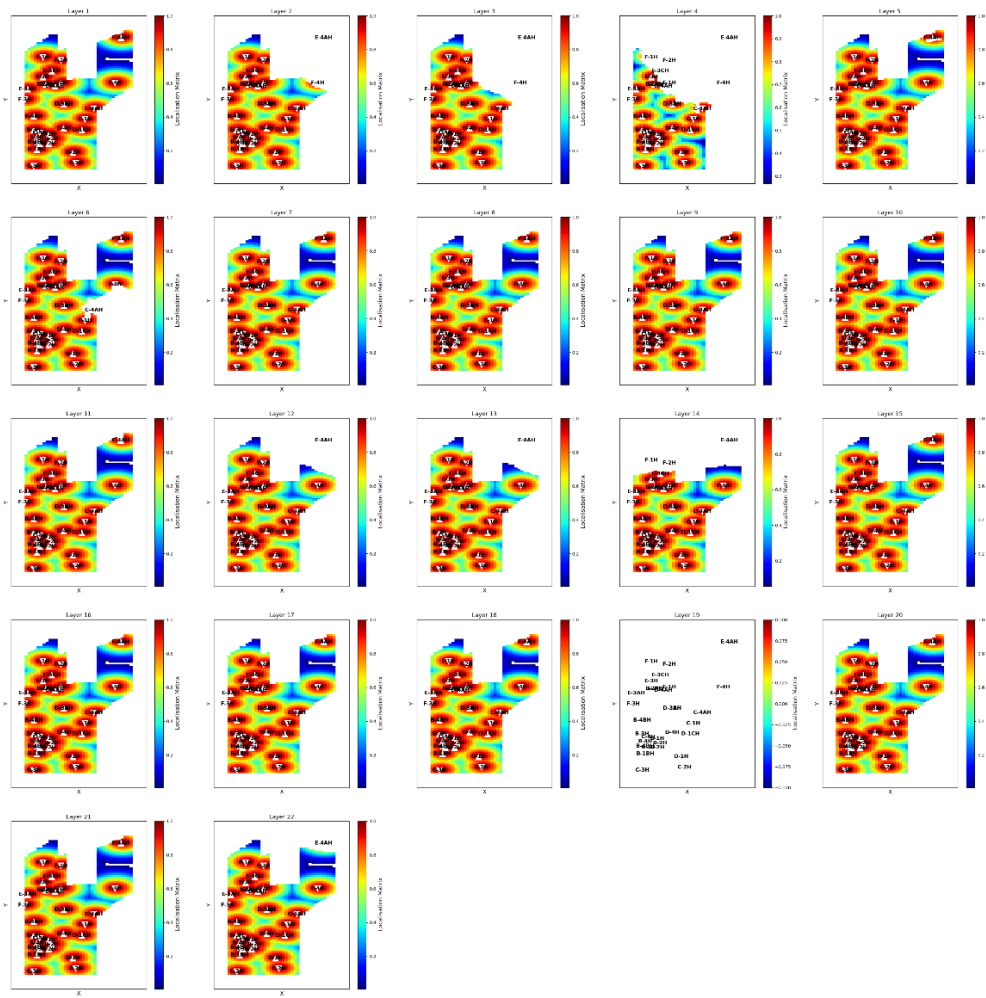

***Figure 27****: Covariance localisation matrix generated from the 5$^{th}$ order compact support Gaspari-Cohn correlation matrix.*

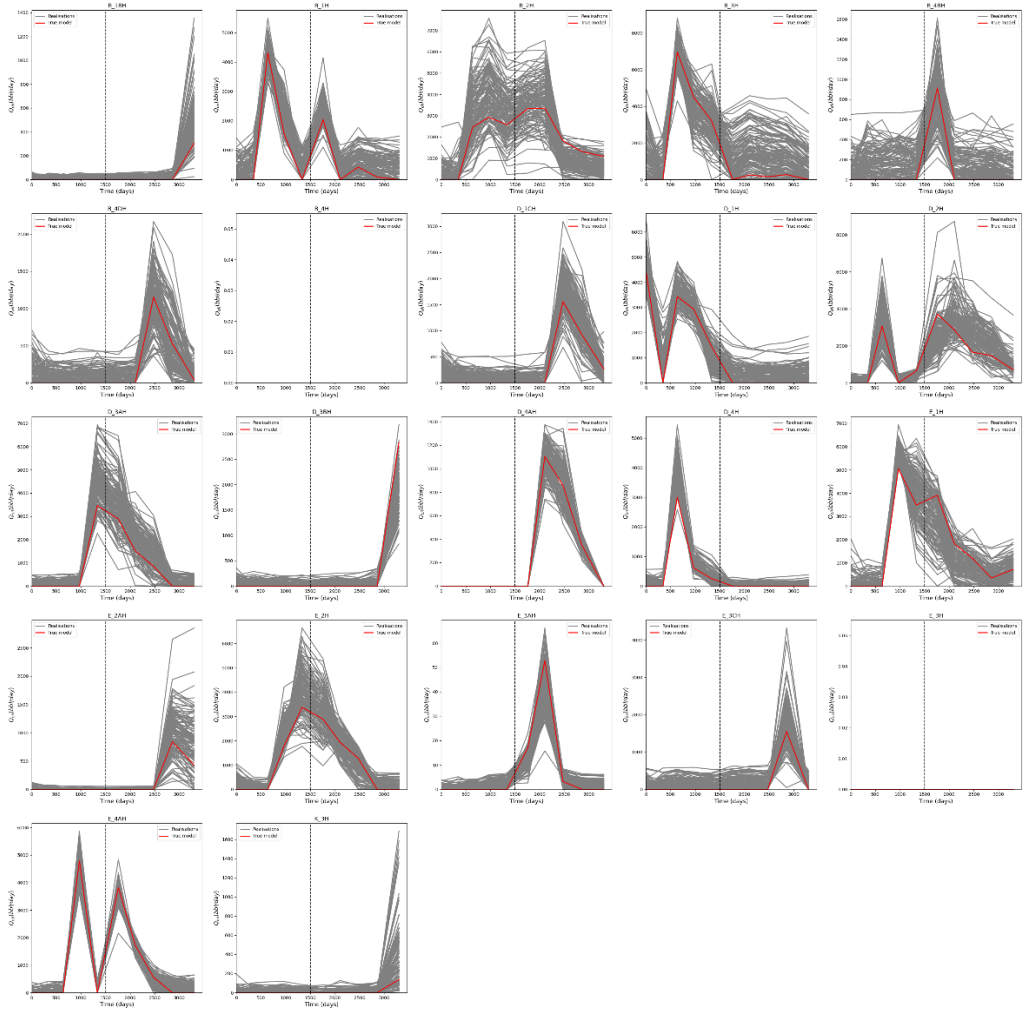

*Figure 28(a)*: *Inverse modelling with covariance localisation and aREKI prior ensemble for oil production rate,* $Q_o \frac{(stb)}{(day)}$

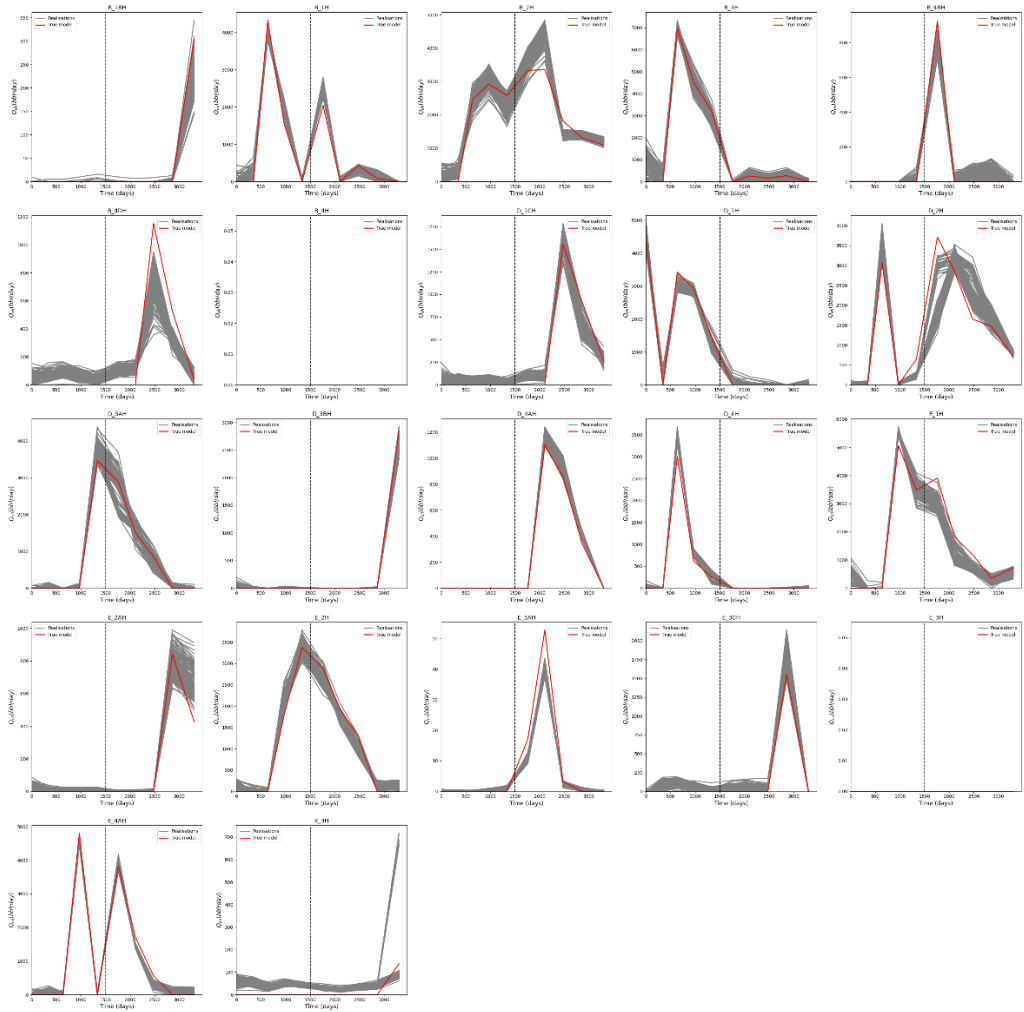

***Figure 28(b)***: *Inverse modelling with covariance localisation and aREKI posterior ensemble for oil production rate $Q_o \frac{(stb)}{(day)}$,*

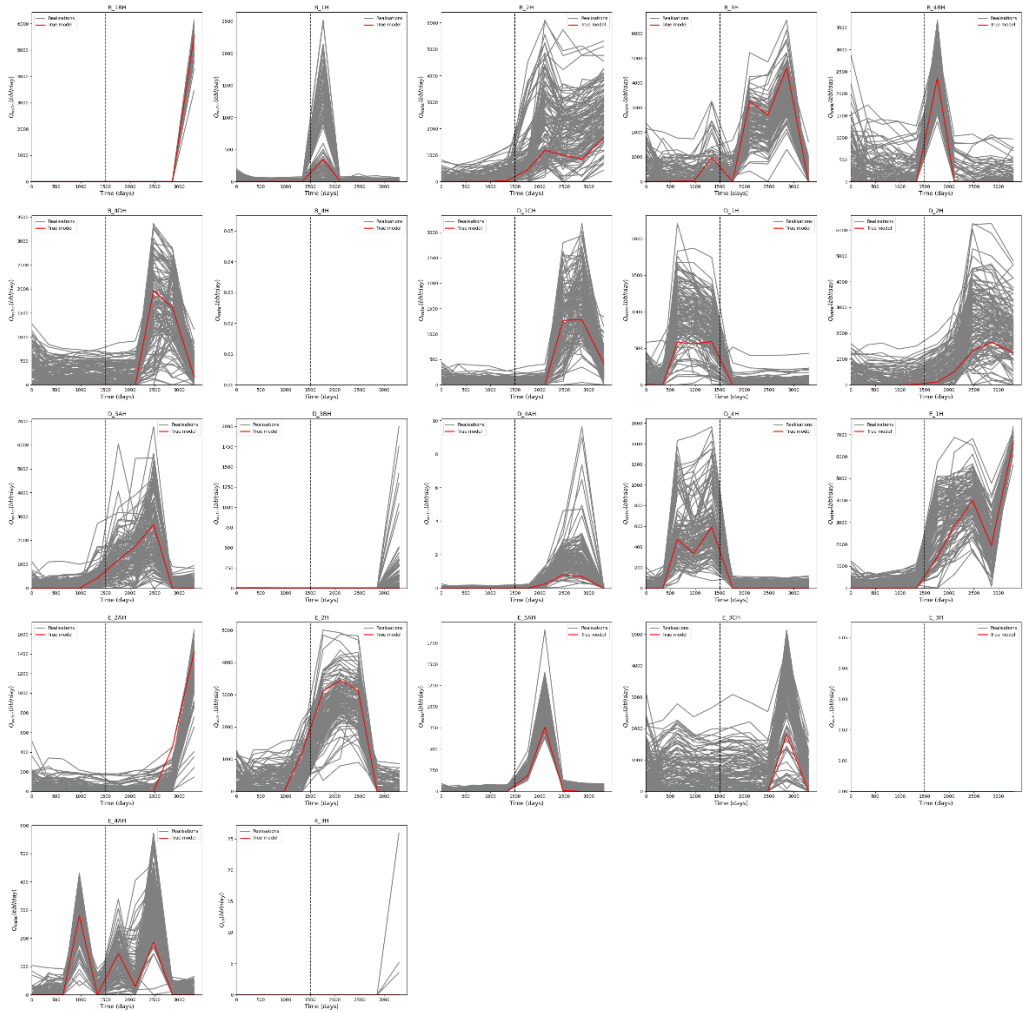

***Figure 28(c)***: *Inverse modelling with covariance localisation and aREKI prior ensemble for water production rate,* $Q_w \frac{(stb)}{(day)}$

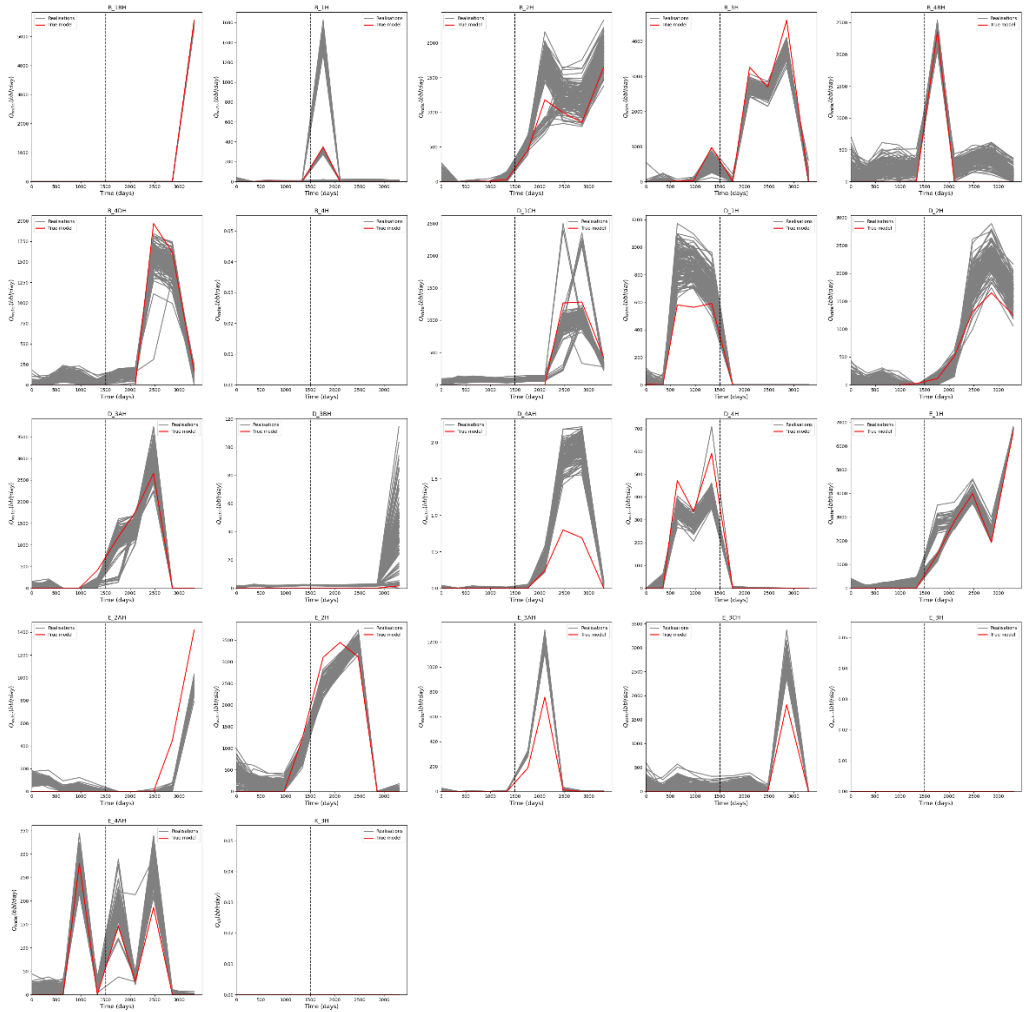

***Figure 28(d)***: *Inverse modelling with covariance localisation and posterior ensemble water production rate,* $Q_w \frac{(stb)}{(day)}$

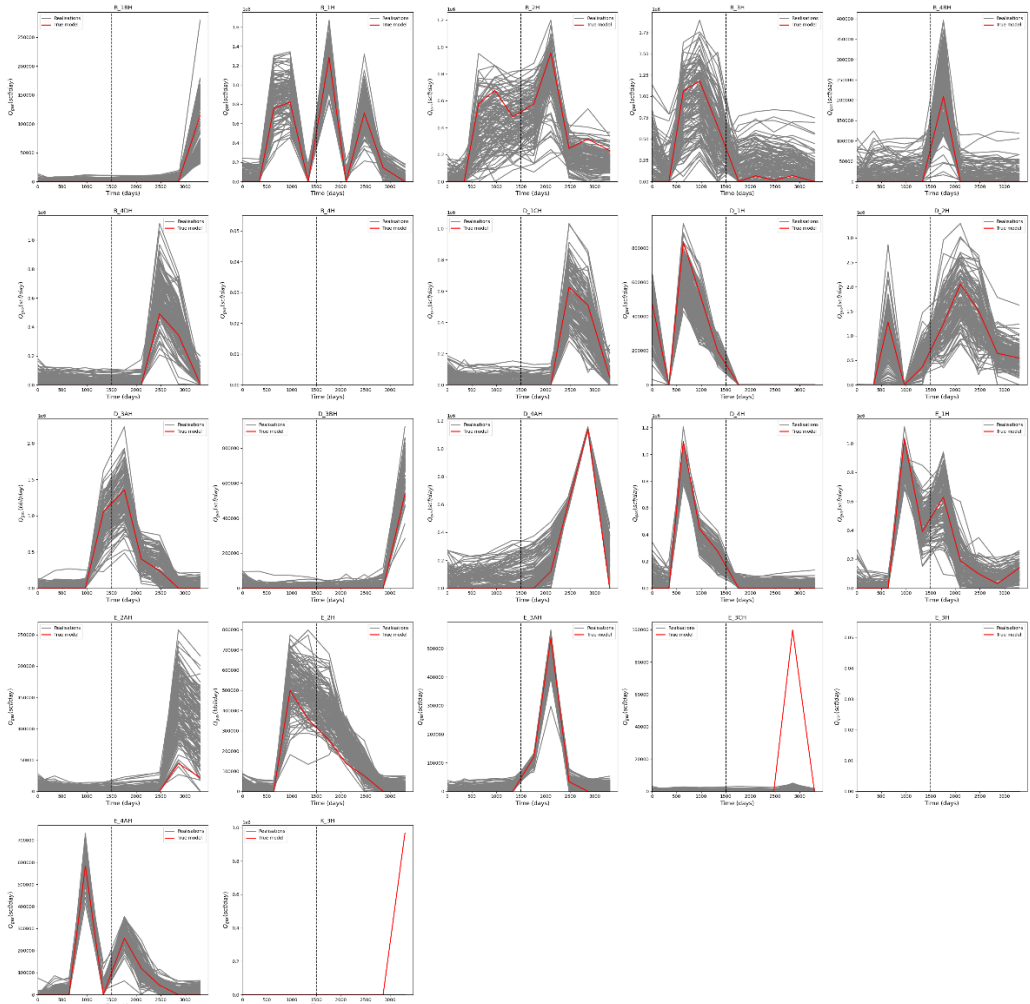

*Figure 28(e)*: *Inverse modelling with covariance localisation and aREKI prior ensemble for gas production rate,* $Q_w \frac{(scf)}{(day)}$.

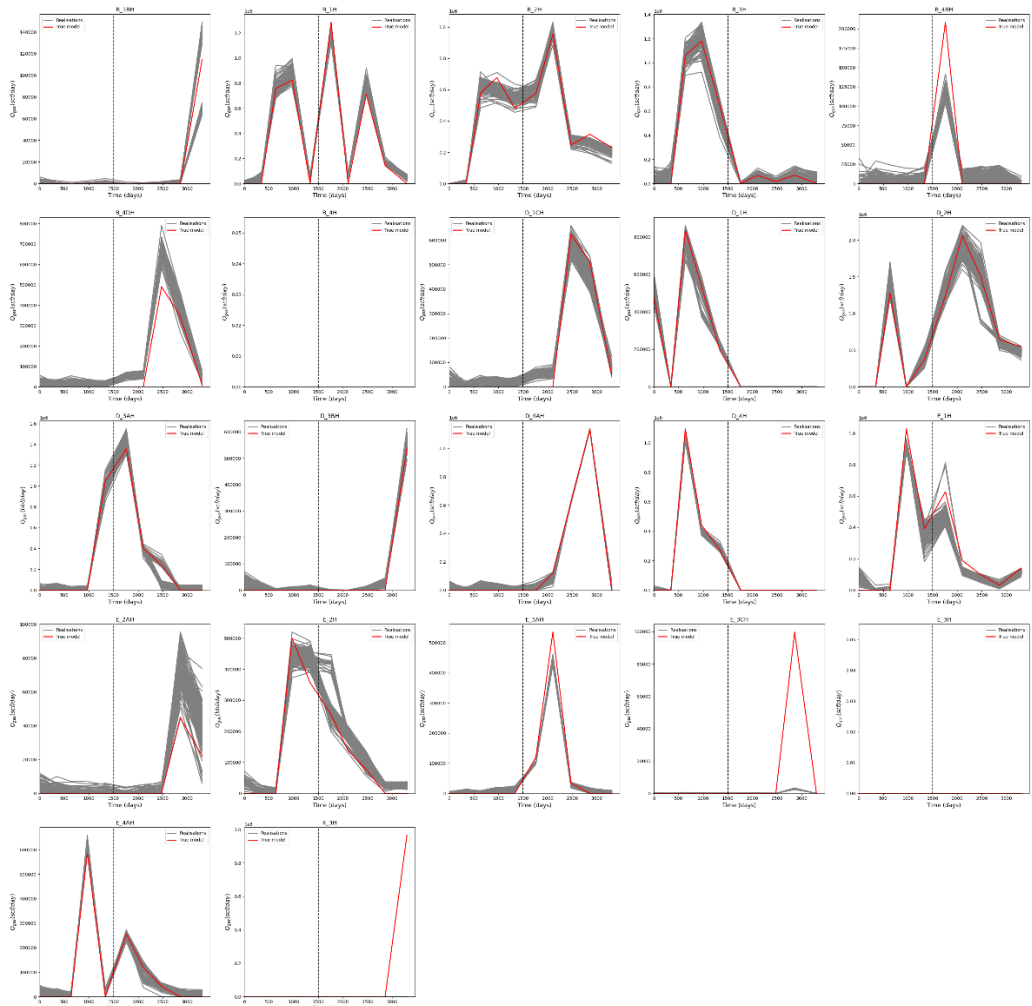

*Figure 28(f)*: *Inverse modelling with covariance localisation and aREKI posterior ensemble for gas production rate,* $Q_w \ \frac{(scf)}{(day)}$

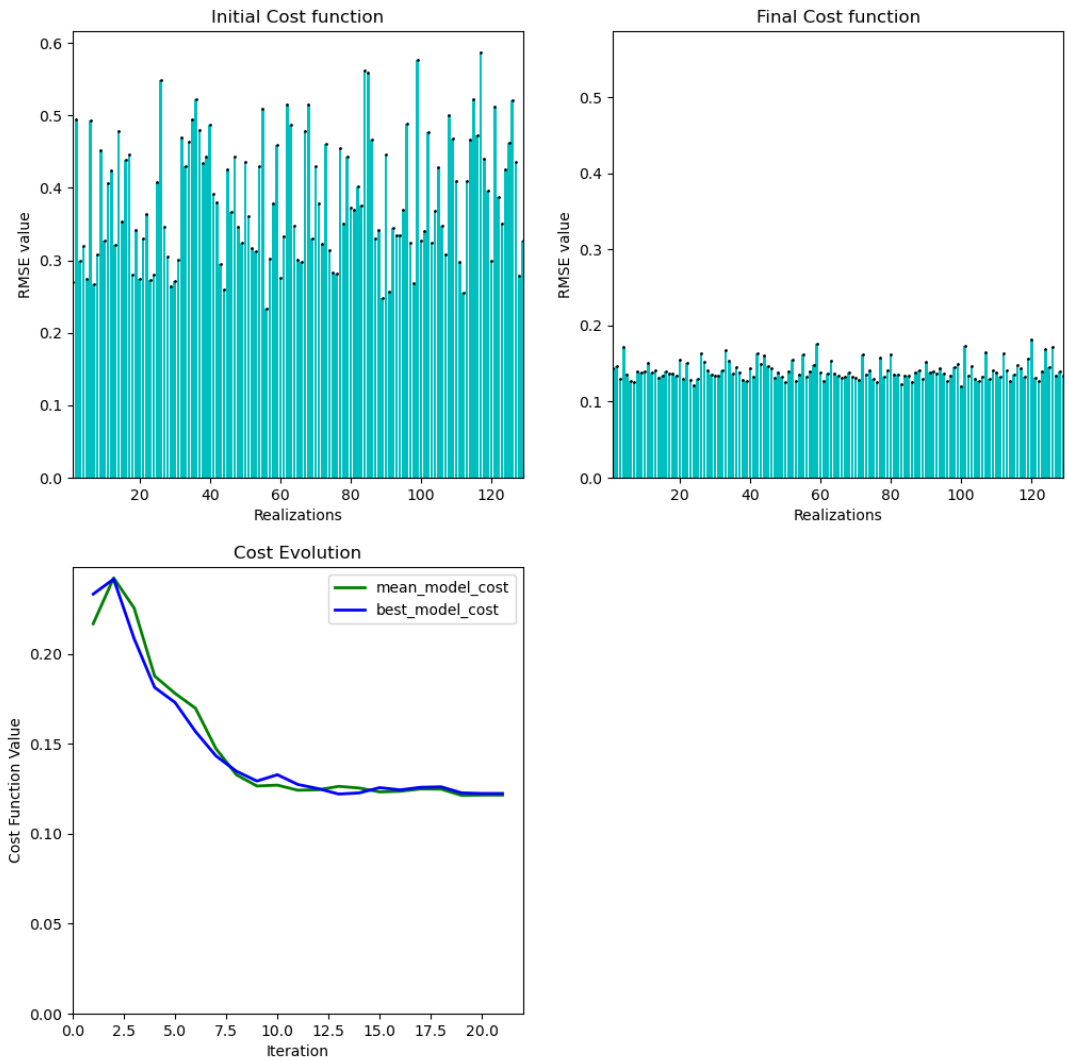

*Figure 29:* *Cost function evolution. (Top-left) prior ensemble RMSE cost, (Top-right) posterior ensemble RMSE cost, (Bottom-left) RMSE cost evolution between the MAP model (blue) and the MLE model (green)*

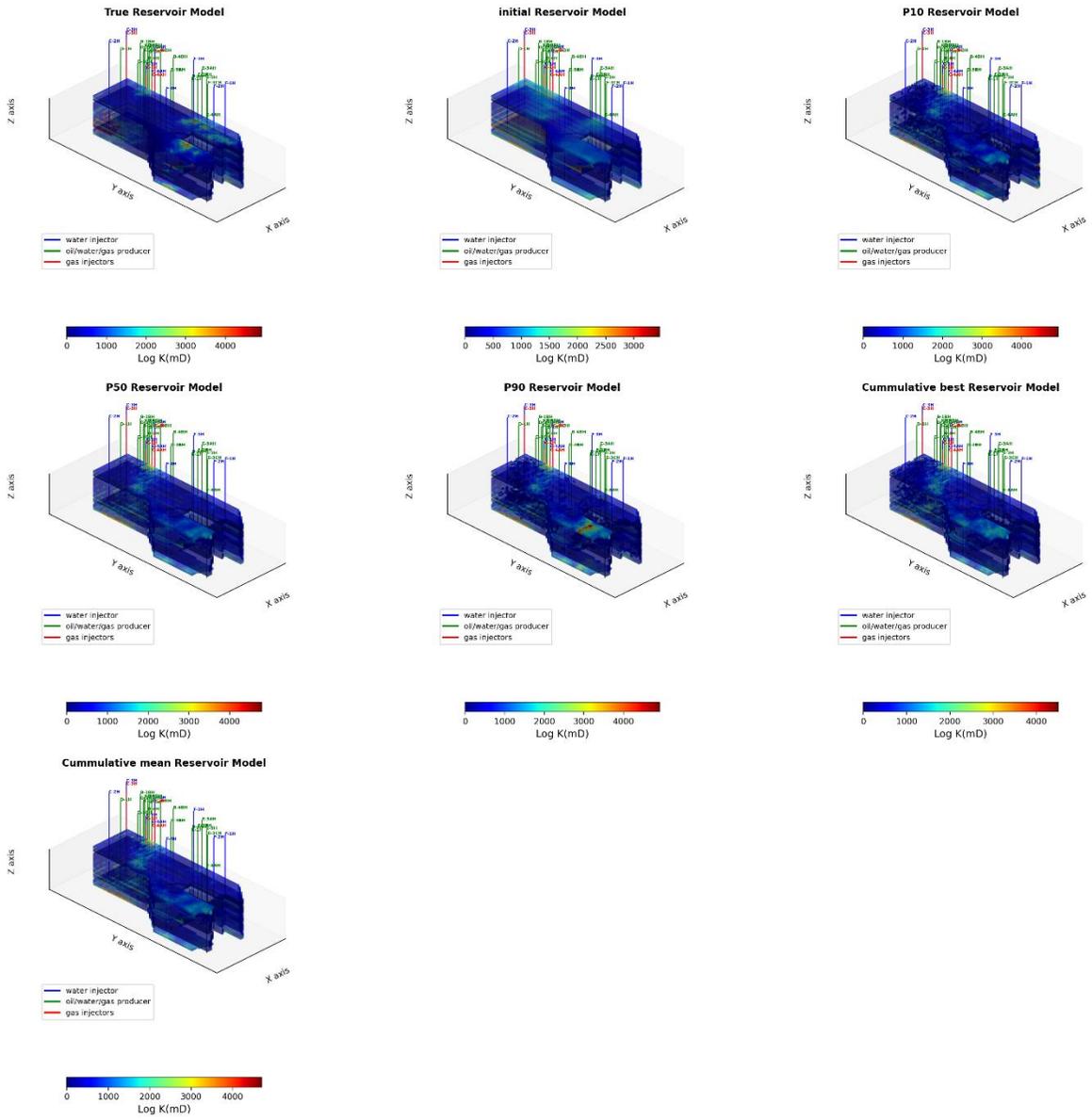

***Figure 30**: Comparison of permeability field reconstruction (1st column) prior, (2nd column) MLE estimate (posterior) (third column) MAP estimate (posterior) and (4th-column) True Model. The method used is the – aREKI + Convolution autoencoder.*

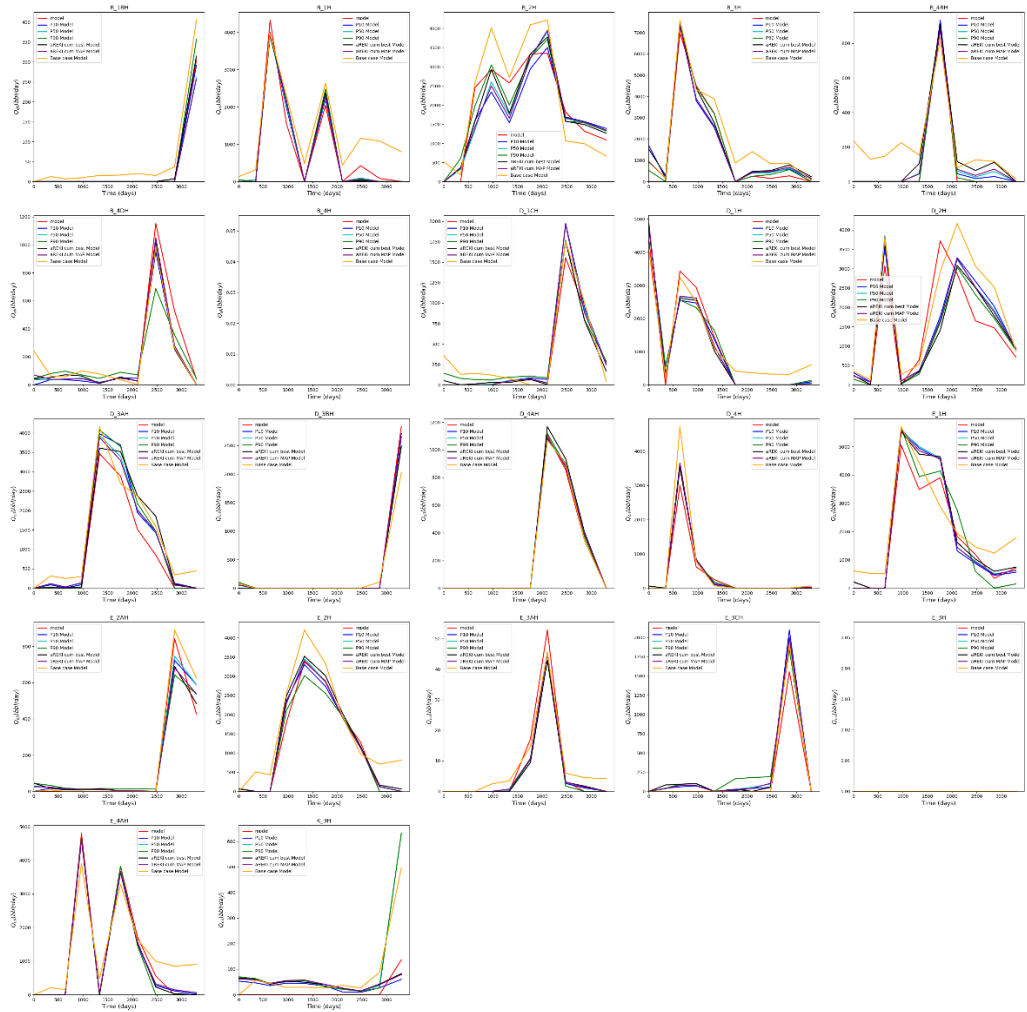

***Figure 31(a)***: *P10-P50-P90 Production profile comparison for the posterior ensemble. (red) True model, (Blue) P10 model, (cyan) P50 model, (green) P90 model. Outputs from peaceman machine for gas production rate,$Q_o \frac{(stb)}{(day)}$ from the 22 producers. They are 22 oil/water/gas producers, 9 water injectors and 4 gas injectors.*

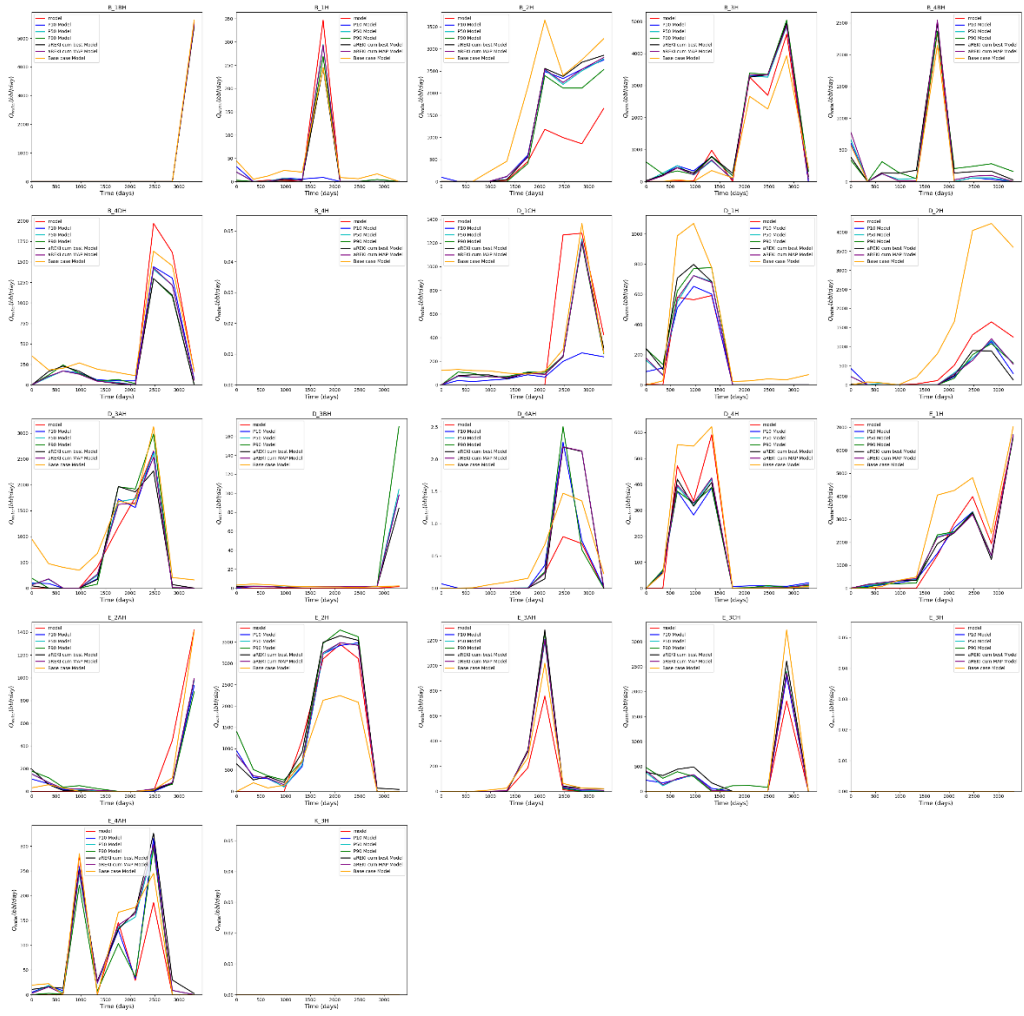

*Figure 31(b)*: *P10-P50-P90 Production profile comparison for the posterior ensemble. (red) True model, (Blue) P10 model, (cyan) P50 model, (green) P90 model. Outputs from peaceman machine for gas production rate,* $Q_w \frac{(stb)}{(day)}$ *from the 22 producers. They are 22 oil/water/gas producers, 9 water injectors and 4 gas injectors.*

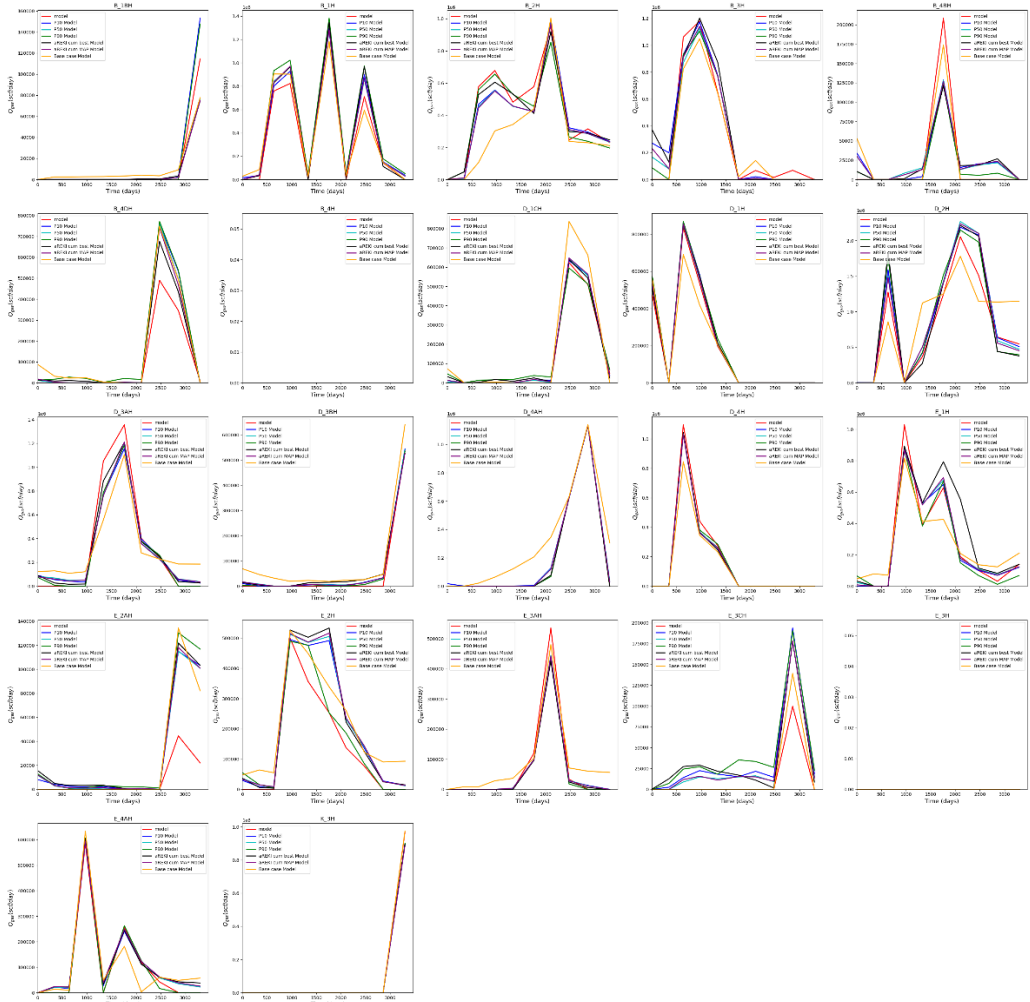

*Figure 31(c)*: *P10-P50-P90 Production profile comparison for the posterior ensemble. (red) True model, (Blue) P10 model, (cyan) P50 model, (green) P90 model. Outputs from peaceman machine for gas production rate,* $Q_g \frac{(scf)}{(day)}$ *from the 22 producers. They are 22 oil/water/gas producers, 9 water injectors and 4 gas injectors.*

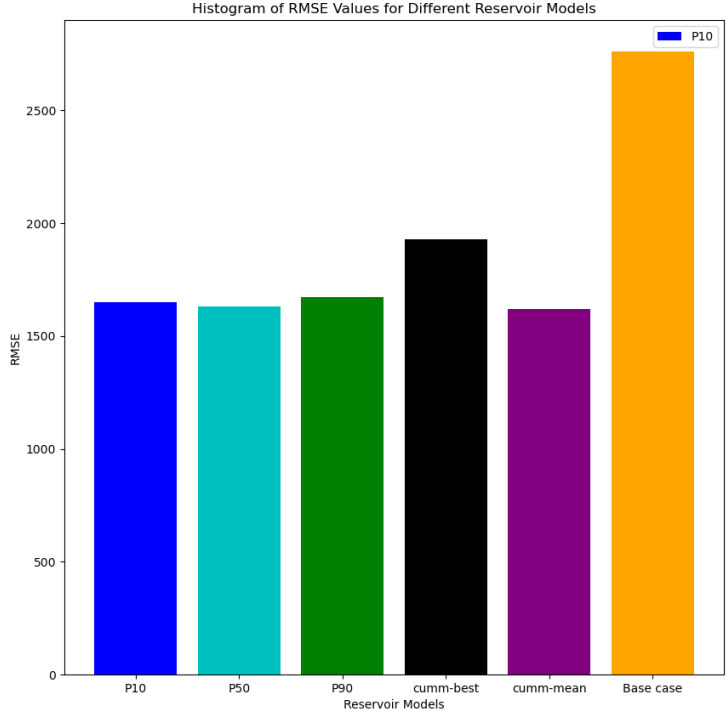

*Figure 32: Inverse modelling of the Norne Field with covariance localisation and aREKI. RMSE value showing the performance of the 6-reservoir model accuracy to the True data to be matched.*

## Conclusion

We have developed a physics-informed neural operator-based reservoir simulator that solves the black oil model by augmenting the supervised loss from the labelled data with the residual loss of the governing pde. The *PINO* can approximate the pressure and water saturation field for any input realization of permeability and porosity for the two-phase problem and approximate the pressure, water saturation and gas saturation given any input of the permeability and porosity field and the fault multiplier term for the three-phase problem. The network is a neural operator. The developed *PINO* is first compared to the numerical black oil reservoir simulator and the similarities are very close. Next, we use this developed surrogate simulator in an ensemble-based history-matching workflow on a synthetic reservoir model. The method used is the *a*REKI with VCAE/CCR parametrization to retain the facies connectivity and sample the non-Gaussian posterior density. The overall workflow is successful in recovering the unknown permeability field and simulation (inference) is very fast with a speedup of 100 - 6000X to the numerical method as we do not need to form the large sparse Jacobian matrix. Training of the *PINO* on a cluster with an NVIDIA H100, takes approximately 30 minutes with 10,000 training samples (9800 samples for the physics loss without any labelling and 200 samples for the data loss requiring labelling and running our in-house numerical simulator). For the *Norne* field, the overall workflow is successful in recovering the unknown permeability field and simulation (inference) is very fast with an inference time of ~7secs as we do not need to form the large sparse Jacobian matrix. Training of the *PINO* surrogate for the *Norne* field on a cluster with 2 NVIDIA H100 GPUs with 80G memory each takes approximately 4 hours for ~100 training samples. Full reservoir characterization together with plotting and solution to the inverse problem for an ensemble size of 200 realizations took ~1 hour(s). The workflow is suitable to be used in an inverse uncertainty quantification scenario where sampling the full posterior density is desirable.

# Credit authorship contribution statement

**Clement Etienam:** Data curation, Conceptualization, Formal analysis, Investigation, Methodology, Software, Validation, Writing – original draft. **Yang Juntao/Oleg Ovcharenko:** Investigation, Writing – review & editing – original draft. **Issam Said/Pavel Dimitrov/Ken Hester:** Research direction, conceptualisation of the forward and inverse problem. **Kaustubh Tangsali:** Technical expertise in the usage of NVIDIA-Modulus-Sym, Writing and editing.

# Acknowledgement

The authors would like to express gratitude to NVIDIA for their financial support and permission to publish the article. Additionally, we extend our thanks to colleagues who contributed to this paper by way of intellectual discussions and suggestions: Tugrul Konuk, Asma Farjallah and Harpreet Sethi.